\documentclass[journal]{IEEEtran}

\IEEEoverridecommandlockouts                              

\usepackage[utf8]{inputenc}
\usepackage{colortbl}
\usepackage{graphicx} 
\usepackage{epsfig} 
\usepackage{epstopdf}
\usepackage{tikz}
\usepackage{mathptmx} 
\usepackage{times} 
\usepackage{amsmath} 
\usepackage{amssymb}  

\usepackage{algorithm}
\usepackage{algorithmic}

\usepackage{tablefootnote}
\usepackage{color}
\usepackage{soul}
\usepackage[normalem]{ulem}
\usepackage{amsfonts}
\usepackage[noadjust]{cite}
\usepackage{url}

\usepackage{threeparttable}

\usepackage{multirow}
\usepackage{array}
\usepackage{float}
\usepackage{multicol}

\usepackage[absolute]{textpos}     

\newcommand{\copyrightstatement}{
    \begin{textblock*}{5.7in}(0.25in,0.5in) 

        \noindent
        \footnotesize
        This accepted article to T-RO 2021 is made available by the authors in compliance with IEEE policy.

        \noindent

        \noindent
        Please find the final, published version in IEEE Xplore.

    \end{textblock*}

    \begin{textblock*}{5.7in}[0,1](0.25in,10.85in) 

        \noindent
        \footnotesize
        \copyright 2021 IEEE. Personal use of this material is permitted.
        Permission from IEEE must be obtained for all other uses, in any current or future media, including reprinting/republishing this material for advertising or promotional purposes, creating new collective works, for resale or redistribution to servers or lists, or reuse of any copyrighted component of this work in other works.

    \end{textblock*}
}

\begin{document}
\copyrightstatement  
\title{\LARGE \bf Quori: A Community-Informed Design of a \\ Socially Interactive Humanoid Robot}

\author{Andrew Specian$^{1}$, Ross Mead$^{2}$, Simon Kim$^{3}$,  Maja Matari\'c$^{4}$ and Mark Yim$^{1}$ %
\thanks{$^{1}$ Andrew Specian and Mark Yim are with the Department of Mechanical Engineering and Applied Mechanics, University of Pennsylvania, Philadelphia, PA, USA.
        {\tt\small aspecian@seas.upenn.edu}%
        \newline
       $^{2}$ Ross Mead is with Semio AI, Inc., Los Angeles, CA, USA.
        {\tt\small ross@semio.ai}%
        \newline
         $^{3}$ Simon Kim is with the the Weitzman School of Design, University of Pennsylvania, Philadelphia, PA, USA.
         {\tt\small simon.kim@i-k-studio.com}%
        \newline
        $^{4}$ Maja Matari\'c is with USC, Los Angeles, CA, USA.
        {\tt\small mataric@usc.edu}}%
}

\markboth{IEEE Transactions on Robotics Regular Paper. December~2020}%
{Specian \MakeLowercase{\textit{et al.}}: Quori}

\maketitle

\begin{abstract}
Hardware platforms for socially interactive robotics can be limited by cost or lack of functionality. This paper presents the overall system---design, hardware, and software---for Quori, a novel, affordable, socially interactive humanoid robot platform for facilitating non-contact human-robot interaction (HRI) research. The design of the system is motivated by feedback sampled from the HRI research community. %
The overall design maintains a balance of affordability and functionality. Initial Quori testing and a six-month deployment are presented. Ten Quori platforms have been awarded to a diverse group of researchers from across the United States to facilitate HRI research to build a community database from a common platform.

\end{abstract}

\section{Introduction}

\label{sec:intro}
\begin{figure}[ht]
\centering
\includegraphics[width=2.7in]{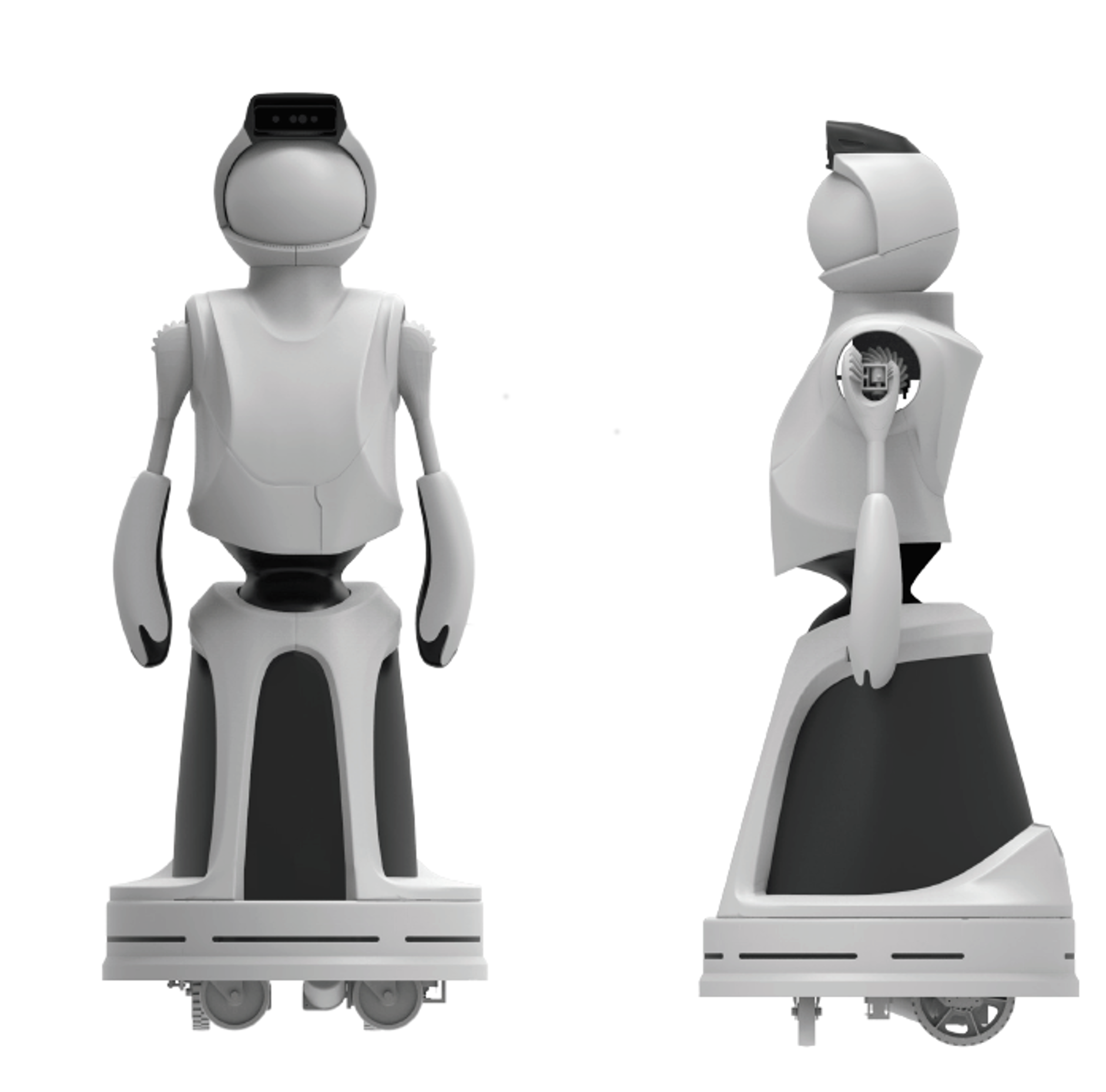}
\caption{Quori's finalized appearance rendered. Photo credit to IK Studio and Immersive Kinematics}
\label{fig:renderComponents}
\end{figure}

This paper presents Quori (Fig.~\ref{fig:renderComponents}), a novel, socially interactive robot for facilitating non-contact human-robot interaction (HRI) research, both in the lab and in the wild. Quori aims to provide an affordable, high-quality platform that allows HRI  researchers to conduct meaningful user studies by deploying systems in the real world. The Quori platform includes both hardware and software to help facilitate HRI. Quori was designed and produced with support from the National Science Foundation Computing Research Infrastructure grant, which included the support for ten Quori platforms to be distributed to researchers in a competitive project proposal process. 

This paper introduces the approach to designing Quori and the details of the complete system. In Section~\ref{sec:Related Work}, we provide some background on HRI. Then, in Section~\ref{sec:designMeth}, we discuss our engagement with the HRI research community that helped inform the most important hardware and software capabilities for a socially interactive robot for HRI research; 
the data collected from this diverse group of researchers in the broader HRI community directed the design decisions for robot hardware and software---this {\em quorum} of researchers inspired the name ``Quori''. The rest of the paper focuses on describing the hardware (Sections~\ref{subsec:appearance}-\ref{subsec:MechDesign}) and software (Section~\ref{sec:Software}). Section~\ref{sec:cost} presents our approach to achieving an affordable design. Section~\ref{sec:testing} describes testing of the system and the four-month deployment of Quori at the Philadelphia Museum of Art.


\section{Background}
\label{sec:Related Work}

HRI research focuses on rich multimodal interaction that occurs between humans and machines \cite{goodrich2008human}. Two subsegments of this field include contact-based HRI and non-contact HRI.

\textit{Contact-based HRI} intersects with medical robotics, haptics, rehabilitation robotics, and is related to manipulation research and to human-robot collaboration.

\textit{Non-contact HRI} grew out of social robotics \cite{breazeal2002designing} and socially assistive robotics \cite{MataricScassselati2016}, and focuses on the perceptual and computational aspects of HRI that involve no physical manipulation of the environment or intentional tactile interactions with people \cite{feil2005defining}, thereby complementing contact-based HRI. Robot hardware and software necessary for pursuing the challenges of non-contact social HRI have many unique requirements. These platforms must be capable of recognizing multimodal social and behavioral signals, reasoning over those signals and behaviors, and generating appropriate affective, expressive, and communicative behaviors in response \cite{breazeal2002designing,feil2005defining}. The range of relevant socially interactive behaviors includes eye gaze (i.e., where/when to look), use of space (e.g.,  where to be, how large to gesture), timing behavior (i.e., turn taking), expressive behaviors (e.g.,  how to gesture, how to move, how to communicate), body language (e.g.,  how to move the body to express a personality/character), and non-verbal and verbal communication (e.g.,  what sounds to use and when, speech processing, natural language understanding, and dialog management), for all of which the problems of behavior recognition, understanding, selection, and control must be solved \cite{mead2014}.

The production of certain social behaviors (e.g., facial expressions and hand/arm gestures) necessitate expressive degrees of freedom (DoFs) on HRI platforms that other robots might not require; for instance, the human face provides rich signaling of affect for human understanding \cite{ekman1999}, so robots with expressive faces can exploit this social communication channel to convey affect and intent.  Expressiveness goes well beyond the face: affect, demeanor, intent, personality, and character are all expressed subtly and effectively through body language. Functional gestures (e.g., deictic pointing gestures) are most readily produced with arms. Therefore, social robots often have arms or other DoFs not used for object manipulation. 

The HRI design features include the robot head, arms, torso and mobility hardware and software and their associated costs.

\subsection{Robot Heads for HRI}
Robot faces can aid in gaze definition, lip readability, and emotional expression \cite{breazeal2002designing} among other purposes that facilitate HRI.  
Faces can be molded and static (e.g., SoftBank's Pepper robot~\cite{pandey2018mass}), can be mechanically actuated for expressivity (e.g., Kismet~\cite{breazeal2002designing} and EMYS~\cite{kkedzierski2013emys}), can be human-like with few DoFs (e.g., Bandit~\cite{fasolamataric2013}) or many DoFs (e.g., Sophia from Hanson Robotics, which use synthetic skin and facial muscle actuators \cite{oh2006design}). Another design solution uses displays for affordable and versatile faces (e.g., Kiwi~\cite{shortetal2017}, which uses Facial Action Units and visemes). Finally, another approach is to use rendered faces projected onto the head, as was done on Quori; as with screens, projected faces have an inherent flexibility for variability and testing, and they also add the potentially more natural head-shaped form, enabling research into how such heads and faces are perceived and accepted by users.

To maximize flexibility of Quori's rendered face, we exploited the availability of affordable portable projectors in a retro-projected animated face (RAF) capable of projecting an image 360$^{\circ}$ around the head. Smaller section RAFs have been shown to be highly expressive, such as in Lighthead \cite{delaunay2009towards}, Mask-Bot \cite{pierce2012mask}, Furhat \cite{moubayed2013furhat}, and Engineered Arts' Socibot~\cite{engineeredarts}. The RAF allows for expressive and customizable faces for HRI research \cite{delaunay2012refined} with potential benefits in eye gaze detection over flat screens \cite{moubayed2013furhat}. 

\subsection{Robot Arms for HRI}
Non-contact HRI robots can use arms for expressivity without touching objects or people. Common functional robot arm gestures include deictic pointing (e.g., \cite{oldpaperbyAaronandRoss}), task demonstration (e.g., in promoting seated exercises for the elderly \cite{fasolamataric2013}), and imitation (e.g., in autism movement training \cite{greczeketal2014}). 
The human arm can be modeled as four DoFs when considering the shoulder and the elbow as described by the model in \cite{tolani1996real}. Since robot cost and complexity increase with the number of DoFs, it is useful to consider functionality with a subset of those DoFs. One DoF can perform simple pointing and express limited affect \cite{misty}.  
Two DoFs in the shoulder allow for more accurate and natural-appearing pointing and expressive motion \cite{brisben2005cosmobot}. Higher DoF arms have been used in rehabilitation and related tasks mentioned above \cite{sobrepera2020design}. Quori's shoulders have two DoFs each, and can modified to add additional DoFs; for details see Section~\ref{subsubsec:ArmDesign}.

Robot arms are an inherent safety risk, due to possible collisions. Safety can be characterized by the Head Injury Criterion (HIC) \cite{zinn2004playing}, a combination of arm inertia and stiffness. %
To ensure a safe low-inertia system, low-mass mechanisms with gravity compensation can be used \cite{whitney2014passively}. Quori uses lightweight arms with a friction clutch and low-power motors to minimize damage from accidental collision. %

\subsection{Torsos for HRI}
Nearly all robots used in HRI research have no flexibility beneath the torso---they lack the ability to bend forward. A waist joint that moves the torso can increase expressivity in a novel way; however, designing a waist joint can be challenging. The weight and motion of components that are not near the axis of rotation require a stiff structure and significant torque to control large torso inertia. Some robot designs avoid the challenges of a waist joint and instead use vertical movement of the torso with a linear actuator (e.g., the PR2~\cite{pr2}). Designs for humanoid robot waists are diverse and range from the hip joint of a legged humanoid robot (e.g., the NAO~\cite{nao}) to commercial wheeled mobile robots (e.g., Pepper~\cite{pandey2018mass} or Car-O-Bot 4~\cite{kittmann2015let}). Two- and three-DoF waist designs use gravity compensation methods, such as the pulley and tendon system \cite{reinecke2020anthropomorphic} \cite{yun20193}. %
A single DoF design allows for simpler gravity compensation; while less movement can be produce by a single DoF, it can produce sagittal motion and still enhance expressivity~\cite{masuda2010motion}.  Quori's one-DoF torso is gravity-compensated, fits within the shell of the robot, and requires no linkages or pulleys.

\subsection{Robot Mobility for HRI}
Some socially interactive robots are \textit{non-mobile}, often operating in tabletop configurations (e.g., Kiwi~\cite{shortetal2017}, LuxAI QT~\cite{luxai}, Furhat~\cite{moubayed2013furhat}, EMYS~\cite{kkedzierski2013emys}, Kaspar~\cite{dautenhahn2009kaspar}, and Keepon~\cite{keepon2014}). 
Most \textit{mobile} socially interactive robots capable use wheels (e.g., Bandit~\cite{mead2015,fasolamataric2013}, Mayfield Kuri \cite{groecheletal2019}, and Pepper~\cite{pandey2018mass}), as they are lower cost, safer, and more easily controlled than legged robots (e.g., Robothespian~\cite{robothes}, Hubo~\cite{park2005mechanical}, and Nao~\cite{nao}).  %
For systems requiring only simple mobility, affordable wheeled systems have employed the Kobuki base, used for the Turtlebot~2~\cite{turtlebot2}, which features a differential drive and a zero-degree turning radius. Alternatively, omnidirectional designs improve the mobility of the system and simplify navigation by removing path-planning constraints \cite{deyle_2010,el2007comparing}. An affordable holonomic mobile base can be achieved with a differential drive mechanism by adding a turret in which its axis of rotation is offset from the midpoint of the two drive wheels \cite{costa2017designing} (Figure \ref{fig:basecad}). This configuration is also called the dual-wheel caster-drive mechanism \cite{wada2000mobile} and is used in the Human Support Robot \cite{yamamoto2018development}, but is otherwise uncommon. %
Quori's holonomic mobile base uses this dual-wheel caster-drive mechanism and is optimized for mobility using the design tools from \cite{costa2017designing}.

\subsection{Affordability for HRI}
The space of robot platforms for HRI research is growing; however, we surveyed the HRI community and found that no HRI robot platform met all the features researchers identified as important: modularity, openness, appearance, actuation, sensing, behavior range, and affordability. The full HRI community survey results are found in Section~\ref{sec:designMeth}.
The PR2 is an open design, modular, and general purpose robot platform that has been highly successful in enabling mobile manipulation robotics research; %
however, it is not designed for social HRI: it lacks social expressiveness and DoFs useful for body language, its hefty size (180 kg) also makes it intimidating for many real-world users, and its cost (US \$400,000) is not affordable to many researchers. 
The NAO from Softbank Robotics has been a popular choice among HRI researchers. It is comparatively affordable (US \$12,000 for one with arms and mobility), and has an appealing design; however, as a closed hardware platform, it is not modular, and  %
it also lacks facial expressiveness. The robot platform that comes closest to meeting the main requirements of the HRI research community is Pepper~\cite{pandey2018mass} (also from Softbank Robotics), which costs of approximately US \$20,000 \cite{pepper_spectrum_2018} (with an annual subscription fee), making it more affordable than the PR2, though more expensive than many tabletop HRI platforms; also, like the NAO, Pepper lacks an expressive face, and features closed hardware and semi-closed software (i.e., closed-source with some open APIs available to interface third-party software) that is robot-specific (i.e., does not work with non-SoftBank robots).

Tabletop platforms offer inherently safe interaction and lower cost, though often in exchange for reduced DoFs and expressivity: some are very simple, such as Keepon~\cite{keepon2014} and Maki (an affordable (US \$5,000 fully assembled) 3D-printed tabletop robot head with mechanical eyes); some are more complex, such as Kiwi (using an adapted Stewart platform for expressive motion and a display for the face ~\cite{shortetal2017}); and some are humanoid, such as Kaspar. Quori's affordability is discussed in Section~\ref{subsec:compare-needs} and  Section~\ref{sec:cost}.%

\subsection{Software for HRI}
Robot control for HRI typically relies on lower-level software to provide standard robotics capabilities, including basic perception and navigation.  Robot Operating System (ROS)\footnote{\url{https://www.ros.org/}} is the most commonly used middleware software platform in general robotics research and is also frequently employed in HRI research. HRI software is then added to handle the various aspects related to the interaction between the robot, the user(s), and the social context. This includes supporting human and activity recognition, speech and natural language understanding, and robot speech, dialog, body language, and facial expression generation. Intuitive tools can aid content creators (who are often non-programmers, including writers and animators) to create and deploy multimodal conversational content for human-robot interactions \cite{mead2017}. HRI systems use standard software tools for voice-based interactions also used in human-computer interaction, including Amazon Alexa, Dialogflow\footnote{\url{https://dialogflow.com}}, and Voiceflow\footnote{\url{https://www.voiceflow.com}}. Analogously, HRI systems use standard software tools from graphics and gaming for animating digital characters, including Maya\footnote{\url{https://www.autodesk.com/products/maya/overview}} and Blender\footnote{\url{https://www.blender.org}}. Because these tools were originally developed for other domains, they typically only offer unimodal capabilities (e.g.,  dialog vs. movement), causing fragmentation in HRI system development.
Our goal for Quori is to provide a more holistic and unified multimodal socially interactive robot software infrastructure capable of supporting face-to-face HRI ``out of the box'' while also being extensible to facilitate HRI research \cite{mead2017}.


\section{Design Methodology}
\label{sec:designMeth}

HRI is a large and rapidly growing field of research and development, involving a very wide range of researcher interests and needs, presenting an exceptional design challenge for a general-purpose socially interactive robot platform.  We employed an iterative community-driven design process to inform the design, hardware, software, and cost of the robot platform. We engaged the broader community of interest through online surveys, conference workshops, and symposia. %

\subsection{Research Community Surveys}
\label{subsec:surveys}
To inform Quori's design, hardware, software, and cost, we distributed two online surveys to the HRI community\footnote{Via mailing lists, such as HRI-Announcement and robotics-worldwide.}. %
Surveys \#1 and \#2 were sent out in Fall 2014 and Fall 2015, respectively, and yielded responses from 37 and 50 survey participants, respectively; nearly all responses were received within the first 48 hours, reflecting community interest and engagement. The surveys elicited considerations regarding (1)~appearance and actuation, (2)~sensing and behaviors, and (3)~cost. The survey results constituted the foundation for the Quori platform design, and are summarized below. Relevant results for Surveys \#1 and \#2 are shown in Appendices~\ref{Appendix:Survey1}~and~\ref{Appendix:Survey2}, respectively; demographics are shown in Appendix~\ref{Appendix:Survey1}, Table~\ref{table:survey1-dem} and Appendix~\ref{Appendix:Survey2}, Table~\ref{table:survey2-dem}, respectively\footnote{Most respondents identified as young and White/Caucasian.}.

\subsubsection{Robot Appearance and Actuation Considerations}
\label{subsubsec:surveyact}
An interactive cartoon-like character with a hard-shell outer covering was preferred over more human-like, biomimetic, or animal-like appearance (Appendix~\ref{Appendix:Survey1}, Table~\ref{table:survey1-app-act}, Prompts~1-2). Survey respondents indicated that the robot should be separated into two independent parts (Appendix~\ref{Appendix:Survey2}, Table~\ref{table:survey2-app-act}, Prompt~4): an expressive upper body, and a mobile base (Appendix~\ref{Appendix:Survey1}, Table~\ref{table:survey1-app-act}, Prompts~6-8~and~9, respectively). The expressive upper body should include actuation of the neck (nodding and shaking), face (eyes, eyelids, eyebrows, and lips), two arms, and if possible a  spine and shoulder actuation (leaning forward and backward, and shrugging) (Appendix~\ref{Appendix:Survey1}, Table~\ref{table:survey1-app-act}, Prompts~6-8 and Appendix~\ref{Appendix:Survey2}, Table~\ref{table:survey2-app-act}, Prompt~2). The mobile base should ideally be omni-directional (Appendix~\ref{Appendix:Survey1}, Table~\ref{table:survey1-app-act}, Prompt~9). Human-robot dialog (through robot speech recognition and production capabilities) is the preferred communication interface (Appendix~\ref{Appendix:Survey1}, Table~\ref{table:survey1-sensing}, Prompts~1-2). The overall robot should be 0.71--1.48 meters in height with the expressive upper body atop the mobile base (Appendix~\ref{Appendix:Survey1}, Table~\ref{table:survey1-app-act}, Prompt~3). Respondents requested the robot be gender-neutral and offered options for establishing a gender identity (Appendix~\ref{Appendix:Survey1}, Table~\ref{table:survey1-app-act}, Prompt~4).

Survey respondents requested ``cartoonish'' physical and social characteristics (Appendix~\ref{Appendix:Survey1}, Table~\ref{table:survey1-app-act}, Prompt~1). In the second survey, respondents did not believe that the mobile base, arms, hands, or chest played a significant role in creating a cartoonish character; instead, they indicated that the use of a retro-projected face, vocal characteristics (e.g.,  speech), and visual behavior (e.g.,  expressive face, arm, and body gestures) would be sufficient for customizing a cartoonish character (Appendix~\ref{Appendix:Survey2}, Table~\ref{table:survey2-app-act}, Prompt~1).

\subsubsection{Robot Sensing and Behavioral Considerations}%
\label{subsubsec:surveysensing}
According to the survey responses, the robot should support both automated perception and control of abilities that are commonly used in face-to-face social interactions (Appendix~\ref{Appendix:Survey1}, Table~\ref{table:survey1-sensing}, Prompts~1-2). The platform should include color (RGB) and depth cameras (RGB+D) for person and object tracking, as well as a microphone array for speech recognition (Appendix~\ref{Appendix:Survey1}, Table~\ref{table:survey1-sensing}, Prompt~3). %
Survey respondents were not consistent with regard to the mobility requirements for human-robot interactions (e.g.,  proxemics), as some researchers preferred that the robot have the ability to move around the environment and others preferred a static tabletop platform (Appendix~\ref{Appendix:Survey2}, Table~\ref{table:survey2-app-act}, Prompt~4), reinforcing our choice to separate the upper body and the mobile base.

\subsubsection{Robot Cost Considerations}
\label{subsubsec:surveycost}
Survey respondents were asked what they would \textit{expect} to pay and what they would be \textit{willing} to pay for a socially interactive robot platform. Those who requested a mobile platform expected to pay \$25,000-\$50,000, while those who requested a tabletop platform expected to pay \$2,500-\$10,000 (Appendix~\ref{Appendix:Survey1}, Table~\ref{table:survey1-cost}, Prompt~1). However, there was high variability in how much researchers were willing to pay (Appendix~\ref{Appendix:Survey1}, Table~\ref{table:survey1-cost}, Prompt~2): the maximum was \$100,000, selected by only 17\% of respondents; 84\% of respondents were willing to pay \$5,000, which we used as the upper bound on the basic Quori hardware platform cost, ensuring that our implementation would meet the needs and budgets of the research community.

\subsection{Community Engagement Meetings}
We presented and discussed Quori prototypes at four research workshops between 2016 and 2018. We hosted two of those workshops: 1) the AAAI 2016 Spring Symposium on ``Enabling Computing Research in Socially Intelligent Human-Robot Interaction: A Community-Driven Modular Research Platform''\footnote{\label{foot:input-meeting}\url{http://www.quori.org/community-input-meetings} (Online proceedings: \url{https://www.aaai.org/Library/Symposia/Spring/ss16-03.php})}, and 2) the Robotics: Science and Systems (RSS) 2016 workshop on ``A Community-Driven Modular Research Platform for Sociable Human-Robot Interaction''\footnote{\url{http://www.quori.org/community-input-meetings/#rss-16-1}}; the other two workshops were: 3) the AAAI 2017 Fall Symposium on ``Artificial Intelligence for Human-Robot Interaction'' \footnote{\url{http://ai-hri.github.io/2017}}, and 4) the 2018 Human-Robot Interaction Conference workshop on ``Social Robotics in the Wild''\footnote{\url{http://socialrobotsinthewild.org}}.

These workshops served to collect community feedback and seek consensus toward Quori's design, with discussions and insights from attendees that complemented the quantitative data we collected with web surveys. At the workshops, we presented a feasibility analysis of Quori's modules \cite{specian2015feasibility} and progress on each module. Attendees of the Spring Symposium were involved in breakout sessions in which they discussed Quori hardware and software design \cite{AAAI_report_2017}, with key feedback from the 2016 workshops strongly informing decisions about the head size and priorities of the arm DoFs. As an indication of active participation, the Symposium included 20 paper presentations$^7$.%
The 2017 and 2018 workshops provided input that helped to finalize the panel design (see next section), as well as determine camera placement. The 2018 workshop provided a means of disseminating Quori's progress \cite{specian2018preliminary} before announcing a call for competitive proposals from researchers interested in participating in the Quori Beta Program; the Call for Proposals was an unexpected source of feedback, in which awardees' desire for a controllable waist DoF justified the additional cost of adding that DoF. The rest of this paper discusses how Quori was designed to address the surveyed and expressed needs of the HRI research community (Fig.~\ref{fig:quoriSensorOverview}). 

\subsection{Comparing Quori to Relevant Platforms}
\label{subsec:compare-needs}
In this section, we compare relevant existing robot platforms in relation to the requirements from the HRI research community presented in Section~\ref{sec:designMeth}. Table~\ref{tab:comp-needs} uses a competitive matrix to highlight the degree to which existing platforms meet the needs we identified in Section~\ref{sec:designMeth}. For this project, we define \textit{open and modular hardware and software} as the ability for the system to be fully observed, modified, and reconfigured (e.g., via adding or exchanging modules) by a researcher; for example, Quori's spine is designed to work with custom arm modules or a head module, and all of Quori's software modules are built using open-source ROS wrappers and communicate via auditable ROS interfaces, so they can be readily exchanged with alternative software implementations. The open-source software, documentation, and hardware designs will be released on the project website (\url{https://www.quori.org)}. The key observation of Table~\ref{tab:comp-needs} is that Quori's design meets the hardware, software, and cost requirements identified in Section~\ref{sec:designMeth}; other relevant HRI platforms either lack affordability or are not as open or modular.

\begin{table}%
\setlength{\tabcolsep}{5pt}
\caption{Existing systems relation to requirements identified in Section~\ref{subsec:surveys}.}%
\begin{tabular}{|m{0.4in}m{0.4in}m{0.4in}m{0.4in}m{0.4in}m{0.4in}|m{0.4in}|}
\hline
\multicolumn{1}{|l|}{}                                                                      & \multicolumn{1}{l|}{PR-2}                                                 & \multicolumn{1}{l|}{iCub}   & \multicolumn{1}{l|}{SociBot} & \multicolumn{1}{l|}{Pepper} & \multicolumn{1}{l|}{Quori}       & \multicolumn{1}{l|}{Kaspar}     \\ \hline
\multicolumn{1}{|l|}{\begin{tabular}[c]{@{}l@{}}Open\\ Hardware\\ /Modularity\end{tabular}} & \multicolumn{1}{l|}{\cellcolor{yellow}Semi}                                                 & \multicolumn{1}{l|}{\cellcolor{green}Yes}    & \multicolumn{1}{l|}{\cellcolor{red}No}      & \multicolumn{1}{l|}{\cellcolor{red}No}     & \multicolumn{1}{l|}{\cellcolor{green}Yes}         & \multicolumn{1}{l|}{\cellcolor{red}No}         \\ \hline
\multicolumn{1}{|l|}{\begin{tabular}[c]{@{}l@{}} Software \\ Open/Closed\end{tabular}}              & \multicolumn{1}{l|}{\cellcolor{green}Open}                                                  & \multicolumn{1}{l|}{\cellcolor{green}Open}    & \multicolumn{1}{l|}{\cellcolor{red}Closed}  & \multicolumn{1}{l|}{\cellcolor{yellow}Semi}   & \multicolumn{1}{l|}{\cellcolor{green}Open}         & \multicolumn{1}{l|}{\cellcolor{yellow}Semi}    \\ \hline
\multicolumn{1}{|l|}{\begin{tabular}[c]{@{}l@{}}Torso \\(Actuation)\end{tabular}}          & \multicolumn{1}{l|}{\cellcolor{green}1-P}                                                  & \multicolumn{1}{l|}{\cellcolor{green}3-R}    & \multicolumn{1}{l|}{\cellcolor{red}0}       & \multicolumn{1}{l|}{\cellcolor{green}2-R}    & \multicolumn{1}{l|}{\cellcolor{green}1-R}         & \multicolumn{1}{l|}{\cellcolor{green}1-R}        \\ \hline
\multicolumn{1}{|l|}{Mobility}                                                              & \multicolumn{1}{l|}{\cellcolor{green}\begin{tabular}[c]{@{}l@{}}12 \\ STLC\end{tabular}}   & \multicolumn{1}{l|}{\cellcolor{yellow}Leg} & \multicolumn{1}{l|}{\cellcolor{red}No}      & \multicolumn{1}{l|}{\cellcolor{green}3 H}    & \multicolumn{1}{l|}{\cellcolor{green}3 H}         & \multicolumn{1}{l|}{\cellcolor{red}No}         \\ \hline
\multicolumn{1}{|l|}{\begin{tabular}[c]{@{}l@{}}Face \\(Expression)\end{tabular}}          & \multicolumn{1}{l|}{\cellcolor{red}Rigid}                                                & \multicolumn{1}{l|}{\cellcolor{green}Mec}   & \multicolumn{1}{l|}{\cellcolor{green}RPF}     & \multicolumn{1}{l|}{\cellcolor{yellow}Rigid}  & \multicolumn{1}{l|}{\cellcolor{green}RPF}         & \multicolumn{1}{l|}{\cellcolor{green}Mec}       \\ \hline
\multicolumn{1}{|l|}{Cost (USD)}                                                            & \multicolumn{1}{l|}{\cellcolor{red}400k}                                                 & \multicolumn{1}{l|}{\cellcolor{red}300k}   & \multicolumn{1}{l|}{\cellcolor{red}30k}     & \multicolumn{1}{l|}{\cellcolor{yellow}20k}    & \multicolumn{1}{l|}{\cellcolor{green}6.4k$^{**}$} & \multicolumn{1}{l|}{\cellcolor{green}2.4k$^{*}$} \\ \hline
\multicolumn{1}{|l|}{\begin{tabular}[c]{@{}l@{}}Size/Height \\(meters)\end{tabular}}                                                           & \multicolumn{1}{l|}{\cellcolor{green}\begin{tabular}[c]{@{}l@{}}1.33-\\ 1.65\end{tabular}} & \multicolumn{1}{l|}{\cellcolor{yellow}1.04}   & \multicolumn{1}{l|}{\cellcolor{green}0.6}     & \multicolumn{1}{l|}{\cellcolor{green}1.2}    & \multicolumn{1}{l|}{\cellcolor{green}1.35}        & \multicolumn{1}{l|}{\cellcolor{yellow}0.55}       \\ \hline
\multicolumn{1}{|l|}{Actuators}                                                             & \multicolumn{1}{l|}{28}                                                   & \multicolumn{1}{l|}{54}     & \multicolumn{1}{l|}{3}       & \multicolumn{1}{l|}{19}     & \multicolumn{1}{l|}{8}           & \multicolumn{1}{l|}{22}         \\ \hline
\multicolumn{7}{l}{\parbox{3.25in}{$*$: Parts cost as reported in \cite{wood2019developing}, $**$: Parts cost without labor and low volume production, H: Holonomic, P: Prismatic Joint, R: Rotational Joint, STLC: Small-time locally controllable, RPF: Rear-projected face, Mec: Mechanical Actuation}}%
\end{tabular}

\label{tab:comp-needs}
\end{table}

%

\section{Physical Appearance}
\label{subsec:appearance}
Physical appearance is a key attribute for a robot designed for social interaction.  We used designer expertise on the team to create an aesthetic that fits the needs of the HRI community as indicated by the community input, allowing modular appearance accessories while retaining stylistic consistency when appropriate. Quori is shown in Fig.~\ref{fig:renderComponents}.

\subsection{Physical Appearance Design}
\label{subsubsec:ApperancedDesign}
Quori has an ``envelope'' design. The underlying robot and mechanical systems are clad with a panelized torso, base, and arms. This allows for the physical mechanisms of the robot and the body shell to remain separated. The design includes ranges of motion for each rigid body part with guarantees of no self-collision. %
Each panel incorporates design considerations for appearance, fit, and finish, as well as ease of disassembly for maintenance and repair. The base panel curvatures are designed to increase the distance between the user and the robot for safety and social proxemics \cite{mead2015}.

Quori's holistic appearance required intentional design. The torso, arms, and base are an identifiable, self-consistent whole: color, seaming, and surface curvature are continuous among the parts. These features address specific community-identified HRI issues of gender, the Uncanny Valley~\cite{mori1970uncanny}, and acceptance. Gender identity is dampened without being generic, the size and appearance are slightly abstract to not mimic human physiognomy and therefore avoid the Uncanny Valley, and the geometry of the robot is meant to be recognizably friendly---we avoided sharp corners and threatening musculature in favor of softly curved surfaces and eased edges that facilitate acceptance. The overall form has large parts with consistent features---the spherical head connects to the rounded torso by a stalk. The out-sized and softly curved forearms meet the torso by a slender femoral shaft. A geometrically simple waist supports the upper body. This yields a perception of a network of discrete, soft spheroids connected by simple masts, rather than a body that is blob-like or mechanical across its surface.

\subsection{Manufacturing and Mechanical Features}
\label{subsubsec:ApperancedMfg}

\begin{figure}[t]
\centering
\includegraphics[width=3.25in]{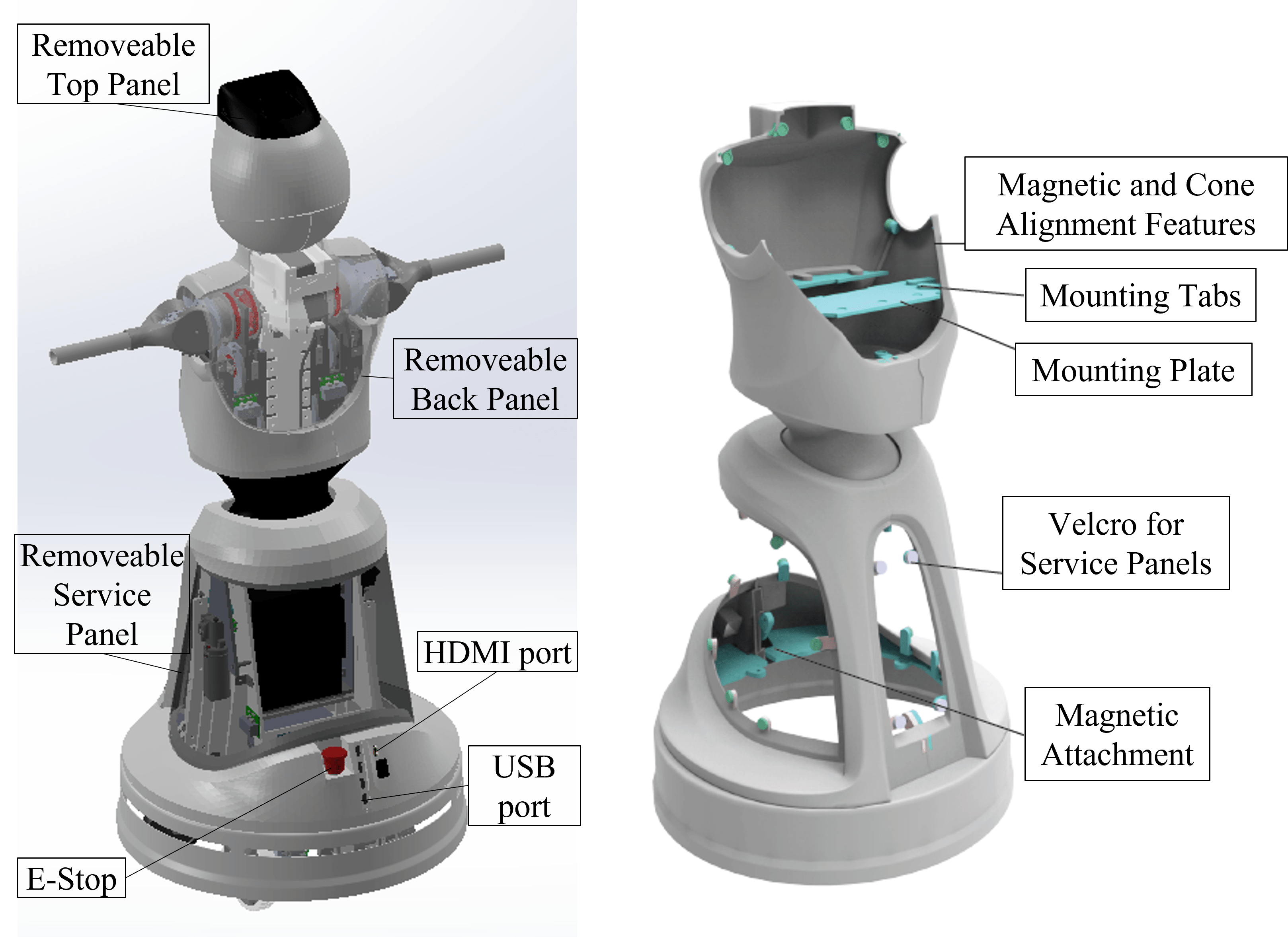}
\caption{Left: Quori has easily removable panels, allowing access to the main computer area, torso, head sensors, and USB and HDMI hub. Right: Design considerations and features for attaching Quori panels. Removable panels not shown to demonstrate ease of access to chest, battery, and computer areas. }
\label{fig:3dprinting}
\end{figure}

As mentioned in Section~\ref{subsubsec:ApperancedDesign}, the panels need to be easily disassembled for repair and robot recharging. There are four panels on the chest (Fig.~\ref{fig:3dprinting}), four constituting the helmet, two for the lower torso (along with two service panel sheets), and one base cover.
The front and back chest panels are removable to allow access to the head projector, the speakers, and the arm controllers, as well as to allow for service or inspection of the upper torso. The helmet parts are removable to allow access to the microphone array and RGB+D camera. The two black service panels (cut from 0.8mm haircell ABS sheets for flexibility, style, and durability) on the lower torso allow for quick access to the main power switch, the battery for charging, and the main computer and its peripheral connections.
3D-printed panels form much of the enclosure. They are interchangeable with different colors or materials, and can be easily removed or replaced via magnetic and mechanical alignment and securing features (Fig.~\ref{fig:3dprinting}, right), avoiding visible mechanical buttons or fasteners on the surface. Panel-mounted neodymium 5mm cylinder magnets with rated 1kg pull force provide enough strength to prevent the panels from shaking apart or falling off while allowing them to be easily pulled off by hand.

A significant amount of labor was required for post-processing. The 3D printed panels required approximately 60 person-hours of in-house labor to improve the finish of the parts, remove printing lines and artifacts, and apply a final color and sealing coat. %
Parts printed with ABS material required light sanding with 400 grit sandpaper to remove printing lines and then painting with a spray primer, followed by white spray-on acrylic gesso, and finally a spray coat of clear varnish. The PLA parts were processed in the same way except for an epoxy coating to fill in the printing layer lines and other artifacts such as glue-seams, since PLA is more difficult to sand than ABS. 
While ABS was easier and faster to process, we preferred PLA parts, as they were easier to print on our printers and significantly less expensive, in some cases nearly half the cost of ABS parts.

\section{Module and Hardware Design}
\label{sec:hardware}

 
\label{subsec:MechDesign}
\begin{figure}[t]
\centering
\includegraphics[width=3.5in]{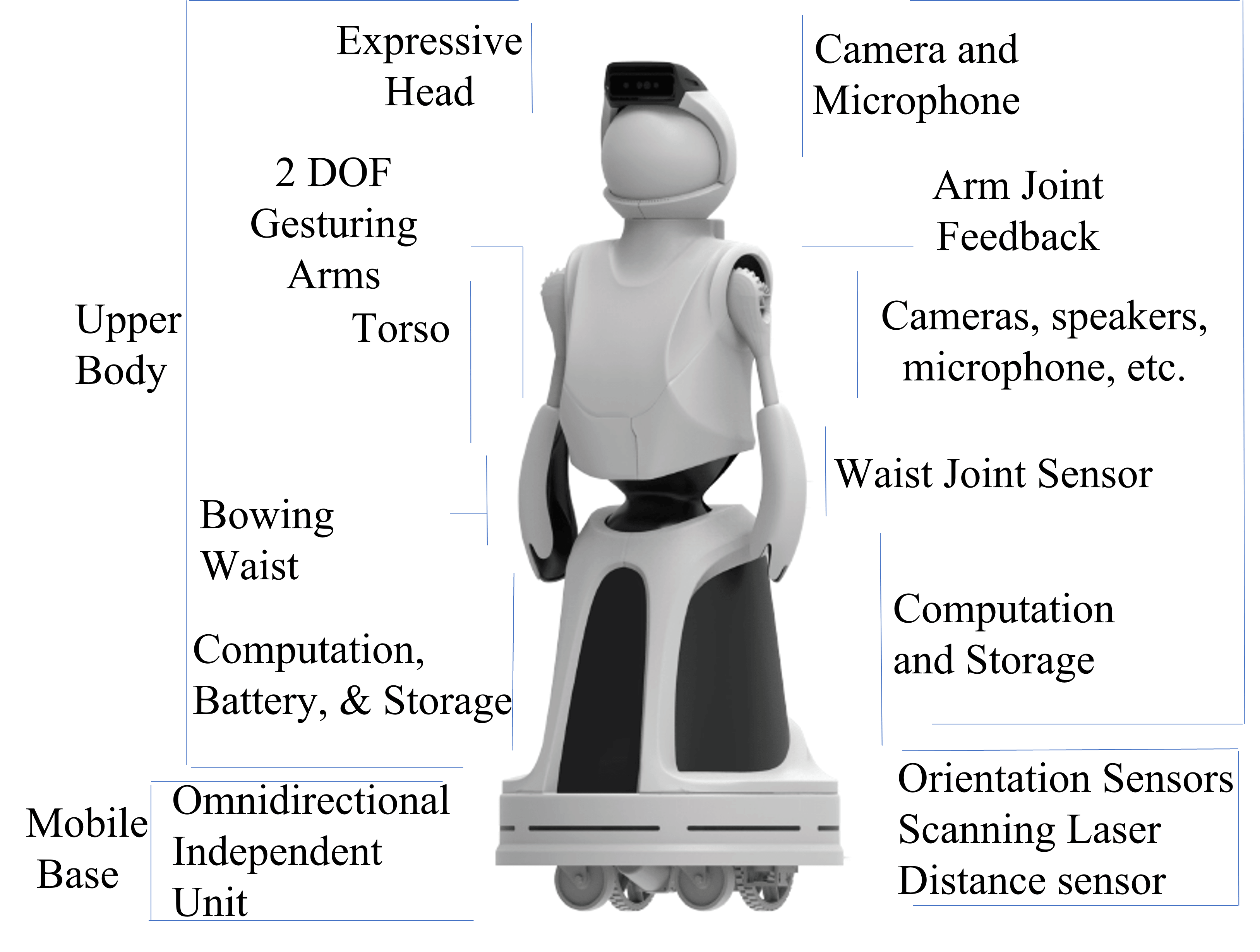}
\caption{Quori's design considerations allow for expressed needs of the HRI community highlighted in the overview of Quori's components (left) and sensing capabilities (right).}
\label{fig:quoriSensorOverview}
\end{figure}

Quori is 1.35m tall, consisting of an expressive upper body attached to a omnidirectional mobile base (Fig.~\ref{fig:quoriSensorOverview}), thus,  within the desired 0.71--1.48 meters in total height (Appendix~\ref{Appendix:Survey1}, Table~\ref{table:survey1-app-act}, Prompt~3). %

Our hardware design consists of three key aspects: (1) validated utility through iterations with the HRI community for desired features,  (2) affordability and targeted feature inclusion, and (3) longevity of impact through development of modular interface standards. The four hardware modules---head, arms, torso, and base---are described in the following sections, along with their power and sensor systems.

\subsection{Cost, Manufacturing, and Design Analysis}
\label{sec:cost}
Our primary mechanism for maintaining low cost was through \textit{elimination} of features. %
By working with the HRI community to identify the most important hardware capabilities for a socially interactive robot, we maximized the value for HRI research while being cost-considerate. The community provided input via online surveys, hosted workshops, and conference presentations (Section~\ref{sec:designMeth}), the feedback for which guided DoF and feature reduction. Three properties not explicitly requested, but kept at high value, were (1) low audible noise from actuators, (2) fluidity of motion, and (3) physical appearance. Manufacturing processes were chosen appropriate for the prototyping quantity: laser cutting, 3D-printing, and water-jet cutting. Parts costs for each module are shown in Table~\ref{table:bom} and sum to US \$6,300.
The highlights of the affordable design feature decisions are:
\begin{itemize}
  \item{\textit{Head:} Elimination of mechanical DoF in the head/neck, using the projected face vs. conventional actuators.} 
  \item{\textit{Arms:} Reduction to two DoF and lightweight arms while enabling modularity and expandability for more DoF.}
  \item{\textit{Torso:} Single DoF with gravity compensation.}
  \item{\textit{Base:} Minimal DoFs for onmidirectional motion with a cost-effective holonomic mobile base drive design.}
\end{itemize}

\noindent Details about each module design are discussed below.
\begin{table}[t]
\caption{Quori's Bill of Materials. Costs are for one unit and do not include savings from bulk purchasing or batch processing. Assembly costs are not included.}
\label{table:bom}
\begin{tabular}{|l|l|r|r|r|}
\hline
\textbf{Subsystem} & \textbf{Item}                     & \multicolumn{1}{l|}{\textbf{Qty}} & \multicolumn{1}{l|}{\textbf{Cost {[}\${]}}} & \multicolumn{1}{l|}{\textbf{Subtotal {[}\${]}}} \\ \hline
Arms {[}2{]}       & Motor modules              & 4                                 & 105                                        & 420                                    \\ \hline
                   & Transmission                      & 2                                 & 170                                        & 340                                    \\ \hline
                   & Joint Sensors                     & 4                                 & 15                                         & 60                                     \\ \hline
                   & Structure                         & 2                                 & 25                                         & 50                                     \\ \hline
Shoulder Joint   &  Joint                    & 2                                 & 67                                         & 134                                    \\ \hline
                   & 3D printed gears                  & 2                                 & 30                                         & 60                                     \\ \hline
Torso              & Motor+Driver+CPU                             & 1                                 & 120                                         & 120                                     \\ \hline
                   & Structure                         & 1                                 & 110                                        & 110                                    \\ \hline
Mobile Base                   & Structure                         & 1                                 & 200                                        & 200                                    \\ \hline
                   & Motor+Driver+CPU                            & 3                                 & 129                                         & 387                                    \\ \hline
                   & Electronics                       & 1                                 & 208                                        & 208                                    \\ \hline
                   & Laser scanner                     & 1                                 & 100                                        & 100                                    \\ \hline
Social Sensors     & Camera                            & 1                                 & 235                                        & 235                                    \\ \hline
                   & Microphone Array                  & 1                                 & 64                                         & 64                                     \\ \hline
                   & Speaker system                    & 1                                 & 18                                         & 18                                     \\ \hline
Head               & Structure                         & 1                                 & 22                                         & 23                                     \\ \hline
                   & Mirror                            & 1                                 & 5                                          & 5                                      \\ \hline
                   & Painted Globe                     & 1                                 & 13                                         & 12                                     \\ \hline
                   & Projector                         & 1                                 & 290                                        & 290                                    \\ \hline
Panels             & 3D Printed & 1                                 & 2000                                       & 2000                                   \\ \hline
                   & Access panels                     & 2                                 & 2                                          & 4                                      \\ \hline
Electronics        & Battery                           & 1                                 & 75                                         & 75                                     \\ \hline
                   & Misc components                   & 1                                 & 305                                        & 305                                    \\ \hline
                   & Onboard computer           & 1                                 & 1100                                       & 1100                                   \\ \hline
                   & \textbf{Total {[}\${]}}                    & \multicolumn{1}{l|}{}             & \multicolumn{1}{l|}{}                      & \textbf{6320}                          \\ \hline
\end{tabular}
\end{table}

%

\newpage
\subsection{Head Design}
\label{subsubsec:HeadDesign}

\begin{figure}[t]
\centering
\includegraphics[width=3in]{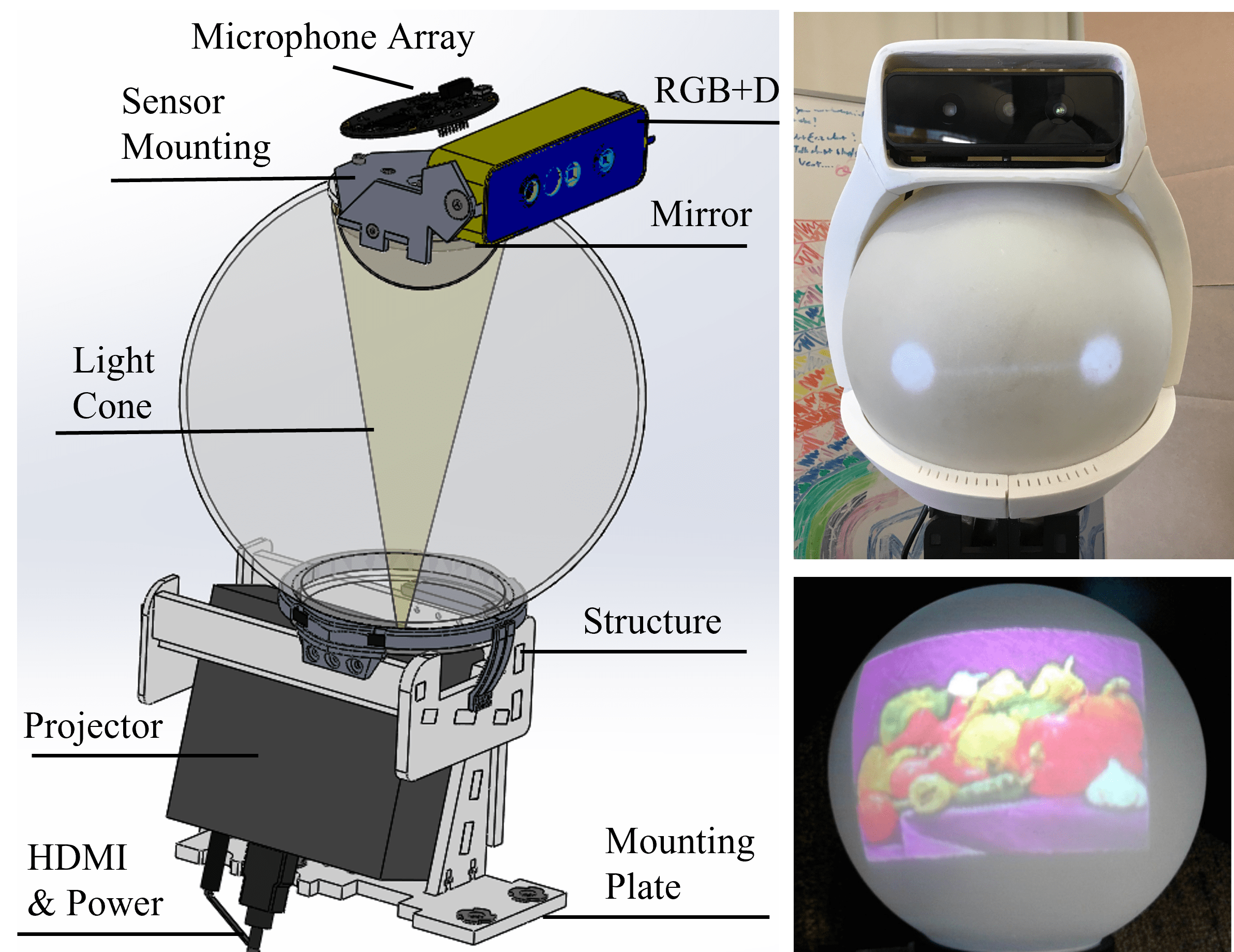}
\caption{Quori's head module is an integrated system that allows for an image to be projected on the surface of the spherical head to produce simple faces (top) or complex images (bottom) via a transformation algorithm~\cite{weng2018low}. The module contains an RGB+D camera mounted directly on the head, and a microphone array mounted on the helmet. The head can be used as a stand alone system.}
\label{fig:headcad}
\end{figure}

Quori's head module uses a retro-projected animated face (RAF) system. It consists of a small projector (115mm x 46mm x 105mm) and a domed mirror to map a projected image onto the inside of specially-coated\footnote{Screen Coating: \url{https://store.gooscreen.com/Rear-Projection_p_27.html}} thin sphere (Fig.~\ref{fig:headcad}). %
These components fit within a compact space approximately the size of an adult human head (200mm diameter) and weigh about 975 grams (excluding the RGB+D camera and microphone array).

The projector is an AXAA P5 (US \$290) with key properties: it is rated to provide 300 lumens, last 20,000 hours, and have native 1280x720 HD resolution; however, only $\sim$132 lumens are available to the spherical surface since the image reflected on the spherical head is a circle inscribed inside the projection rectangle. The projector is affordable (at US\$290), has intermediate brightness, and short throw (20cm creates a focused 7.5cm x 12.7cm image). It creates a color image that is visible in most illuminated indoor environments where there is no sunlight saturation (Fig.~\ref{fig:headcad}, top). %

The mapping of projected images onto the sphere's surface is not uniform---the resolution is dense near the top of the head and sparse near the neck. The least dense equatorial line is approximately 200 pixels, compared the highest ring, which has over 2,000 pixels; thus, creating expressive faces to be displayed on the spherical surface via a projected image is not trivial. Our mapping algorithm transforms pixels on the sphere to pixels in a 2D image to be sent to the projector. Details of the design and mapping used to project images to Quori's head can be found in our previous work \cite{weng2018low}. 

The illusion of motion (e.g., head shaking, nodding, and gaze directing) can be produced through projection. Since the robot's head is a rotationally symmetric sphere with no protruding features (e.g., no nose or ears), head rotation can be simulated by projecting the image of a face rotating on the sphere without requiring additional motors or neck DoFs. Gaze direction can be simulated by coupling animation of the eyes with horizontal rotation of the whole upper torso ($M_T$ in Fig.~\ref{fig:basecad}). The waist joint (Fig.~\ref{fig:torsocad}) may also be useful in supplementing gaze, especially for interactions below or above the neutral gaze of the robot (e.g., for users who are shorter or taller than the robot) or for objects very near or far away. Sensors in the head can be replaced via fasteners on the sensor mounting plate. The camera field of view is discussed in Section~\ref{subsec:PowSysDesign} and shown in Fig.~\ref{fig:camerafov}.%

%

\subsection{Arm Design}
\label{subsubsec:ArmDesign}

\begin{figure*}[t]
\centering
\includegraphics[width=\textwidth]{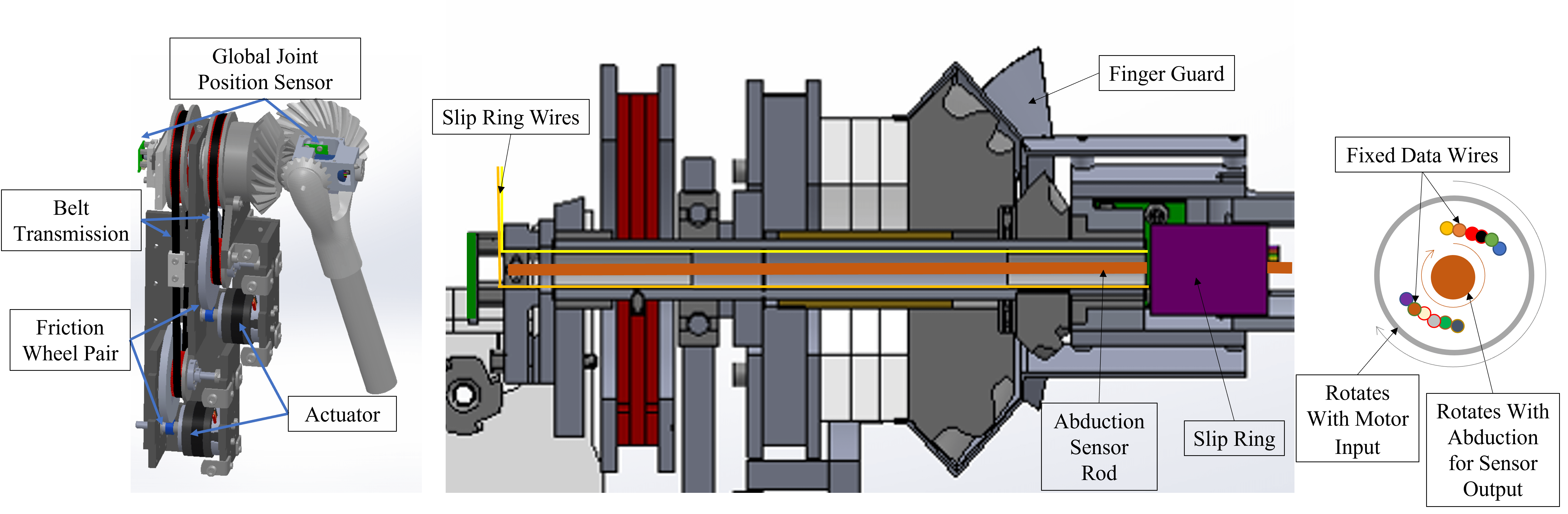}
\caption{Left: CAD model of the arm module. Center: A sectional view of the compact differential transmission.  Right: Section view of the arm differential highlighting how the torque is transferred while allowing 12 wires to be available to the arm with continual shoulder rotation}
\label{fig:armcad}
\end{figure*}

Gestures are a key part of natural communication in social interaction. Quori's arm design is affordable, modular, safe, and expandable. The shoulder module (Fig.~\ref{fig:armcad}) has two DoFs based on a design by \cite{whitney2014passively}; however, our design differs in the use of 3D-printed bevel gears instead of a capstan cable drive. In addition, to save costs and complexity, we chose to \textit{not} gravity-compensate the arm, thus, enabling the elbow and arm modules to be changed. The arm is driven by brushless DC motors through a transmission consisting of a friction wheel pair and a timing belt speed reduction (Fig.~\ref{fig:armcad}, left). The entire arm module mounts to the spine with fasteners. %

Notable features of the arm design include resolution of the joint positions, drive motor abilities, and general safety considerations. The approximate resolutions of the joint position sensors are 0.022$^{\circ}$ and 0.075$^{\circ}$ for the shoulder joint (through the use of magnetic encoders on the output shafts, Fig.~\ref{fig:armcad}) and the drive motors, respectively. Access to both motor and shoulder positions allows the system to check for slippage between the friction wheel pair or timing belt stages, as well as perform automatic calibration upon boot-up of the system.
The arm motors can produce approximately 0.15Nm and are able to rotate at approximately 16 revolutions per second, resulting in  shoulder joint speeds up to $\sim$1.2 radians per second\footnote{This is based on the motor properties at 12 Volts operation.}.
The abduction/adduction DoF has a range of $\pm70^{\circ}$, and the circumduction DoF is continuous. The arm design is expandable; we designed access for power and/or communication for further joints in the arms (e.g., an elbow), while allowing the arm to rotate continuously. We achieved this via a shoulder joint slip ring with six available wires (rated to 2 A) (Fig.~\ref{fig:armcad}, right). As an example, two servo motors can be added as they only use three wires each. More DoF may be added through multiplexing. %

We used the following operational safety measures: a torque limit on the drive motors; a low-mass, low-inertia arm mechanism and structure that is safe according to the Head Injury Criterion \cite{zinn2004playing}; and a friction wheel designed to slip in case the motor generates too much torque or the arm is back-driven. %

Our primary goals in arm design were to ensure safe and precise yet fluid motion for expressivity, while maintaining affordability. Manipulation (i.e., carrying some payload or applying forces to the environment) was explicitly \textit{not} the goal of our design; thus, we used light-weight limbs and IQ Control's position controlled, direct drive, and brushless servo motors\footnote{http://iq-control.com}. Arms that would be expected to lift, push, or pull would need structural stability that typically leads to heavier and more expensive designs. Furthermore, heavier arms require larger and thus more expensive motors to move. Lower-cost motors or servos could be used at the expense of precision for the case of brushless DC motors \cite{piccoli2016anticogging}.

%

\subsection{Torso and Waist Design}
\label{subsubsec:TorsoDesign}
\begin{figure}[t]
\centering
\includegraphics[width=3.25in]{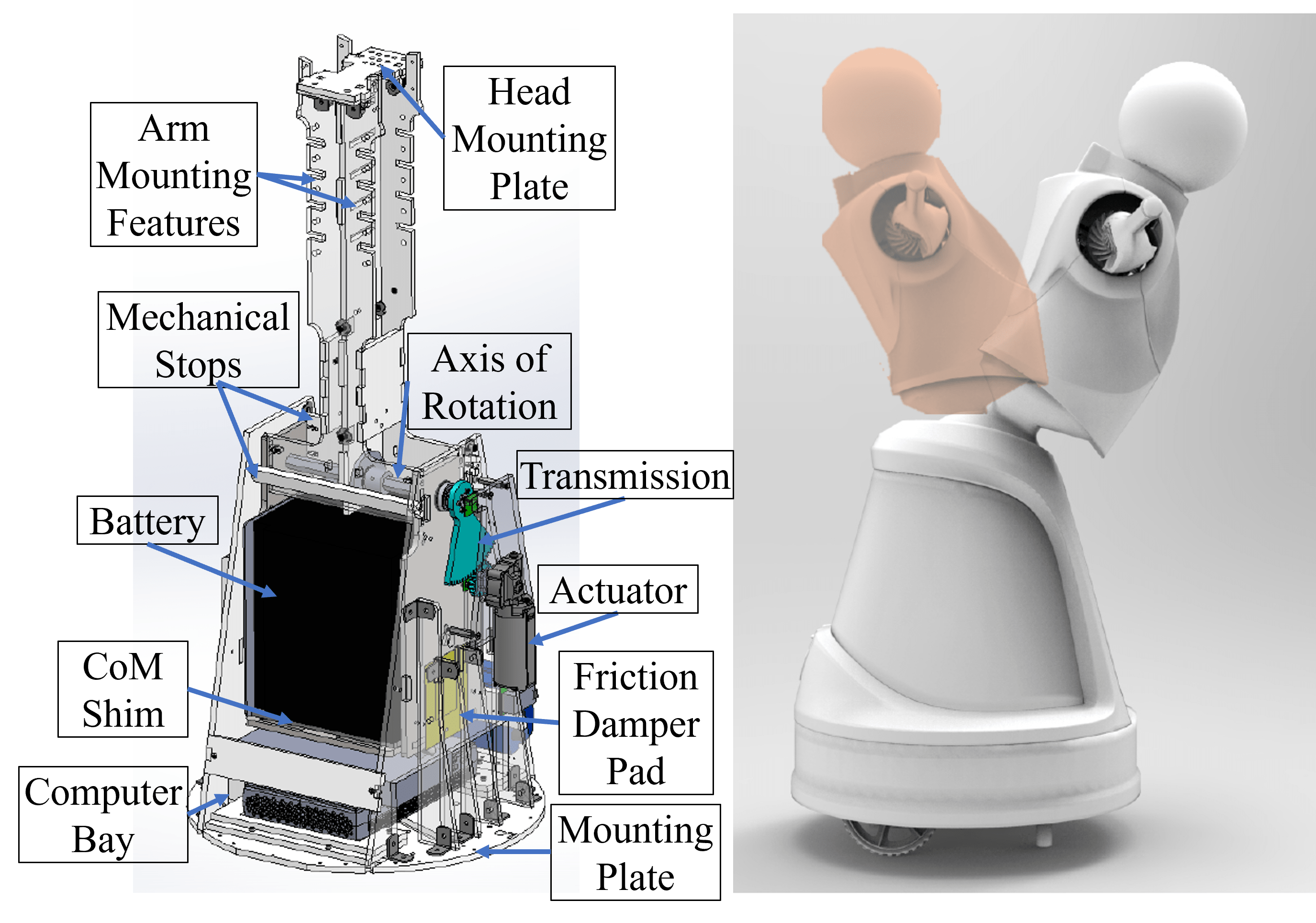}
\caption{Left: Upper torso and waist hardware overview. Right: Extreme positions the robot achieves by bowing forward 30 degrees or leaning back 15 degrees.  Mechanical limitations on the positions prevent self-collision.}
\label{fig:torsocad}
\end{figure}

Quori's torso module not only supports the arms and head (Fig.~\ref{fig:torsocad}, left), but also has one DoF to lean forward and backward (Fig.~\ref{fig:torsocad}, right). %
This design also minimizes acceleration-induced swaying generated during the motion of the mobile base, leading to fluid, natural, and appealing tunable motion. %
The batteries and onboard computer are stored in the torso.%

The spine allows for easy attachment of additional custom hardware, such as arms or a head. A new head module can be attached to the spine using the provided mounting holes. The arms have similar mounting possibilities---shelves/ledges can be added to the spine for additional accessories, such as sensors, tablets, trays, container mountings, etc.

Considerable space is allotted for batteries and a computer\footnote{Quori currently ships with a nuc8i7hvk: Intel$^\copyright$ Core$^{\text{TM}}$ i7-8809G Processor with Radeon$^{\text{TM}}$ RX Vega M GH graphics (8M Cache, up to 4.20 GHz)}:  17cm x 15cm x 21cm and 20cm x 20cm x 7cm, respectively. Currently, the battery bay space fits a 40-ampere-hour sealed lead acid battery that powers the whole robot. While many options exist for small form factor computers, we have ensured sufficient space for a computer with computational resources suitable for real-world use, such as a NUC\footnote{\url{https://ark.intel.com/content/www/us/en/ark/products/126143/intel-nuc-kit-nuc8i7hvk.html}}  or NVIDIA Jetson TX1\footnote{\url{https://developer.nvidia.com/embedded/jetson-tx1-developer-kit}}.

Next, we present our approach to designing the single DoF waist which involves gravity compensation. The transmission design is also discussed.

\subsubsection{Gravity Compensation Design and Tuning of Waist Joint}
Robot motion is often caricatured as jerky with overshoot; for example, a person pretending to be a robot might exaggerate leaning backwards as they start to walk forward, then sway forward and back just as they stop walking, as a cantilevered stick might do as a damped oscillator. Avoiding these types of motions typically requires expensive, strong motors and precise feedback. %
Alternatively, adding mass to shift the center of mass (CoM) can change this behavior. A CoM below the axis of rotation causes the torso to lean forward during acceleration (opposite to the prototypical robot caricature motion), while a CoM at the axis of rotation reduces the motion.  %

\begin{figure}[t]
\centering
\includegraphics[width=3.25in]{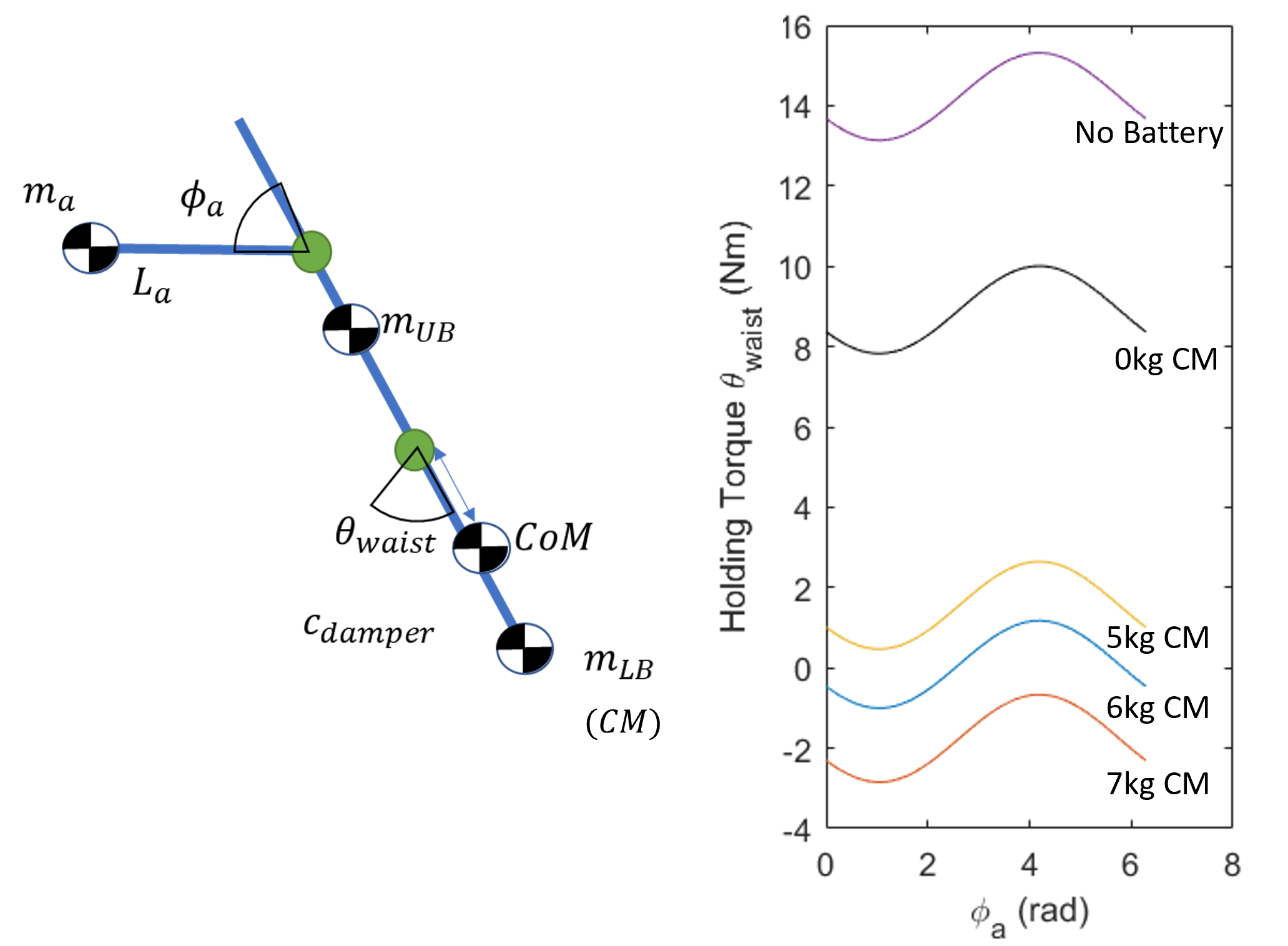}
\setlength{\unitlength}{1in}
\begin{picture}(0,0)
\textbf{\put(-1.5,2.5){\scriptsize Holding Torque as a Function
}
\put(-1.3,2.38){\scriptsize of Flexion Arm Position
}}
\end{picture}
\caption{Left: Model used for tuning the CoM of the torso and waist actuator torque. The masses are separated into the upper body mass $m_{UB}$ (head and arm transmission), the arm link mass $m_a$, and the lower body mass $m_{LB}$ (battery and counter masses).  Right: Maximum waist holding torque curves used to select a starting counter mass (CM) configuration for the torso. The lines are produced by simulating the arms flexion, $\phi_a$, in order to produce the maximum waist torque, $\Ddot{\theta}_{waist}$, to hold the most difficult position of bowing forward. }
\label{fig:comstudy}
\end{figure}
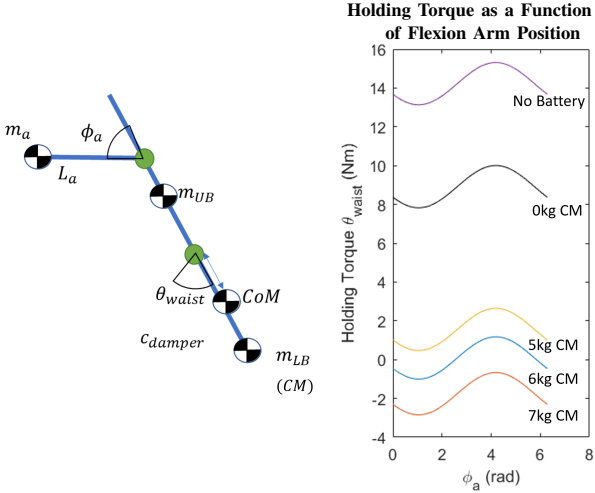

Affordable actuation of the waist can be achieved with a counterbalance metronome design (Fig.~\ref{fig:comstudy}, left). This design leverages the mass of the robot's battery to offset the moment of the upper body of the torso, head, sensors, and arms. The moment that needs to be balanced changes, as the balance depends on the position of the arms $\phi_a$ which may be moving. Fig.~\ref{fig:comstudy}  shows the  torque required to hold the torso at its max bow position as the arms rotate in the plane. The effect of the extra counter-mass, the battery, and an ideal tuned counterbalance design is shown in Fig.~\ref{fig:comstudy}. In its most difficult bowing position, the waist experiences a 16-Nm moment without counterbalancing (Fig.~\ref{fig:comstudy}, purple line). It is very challenging to find a motor with this capability, that is also small enough to fit in the required space and is affordable. Instead, with proper counterbalancing, the battery and an extra 6 kg (Fig.~\ref{fig:comstudy}, blue line), the peak torque requires less than 2-Nm. %
The major drawback to this counterbalance design is the increased inertia of the torso. The waist does not need to move very fast---less than  1$\frac{rad}{s}$---nor accelerate faster than 1$\frac{rad}{s^2}$, which leads to a max accelerating torque of about 2.5 Nm; for reference, the max required static holding torque of the final design is approximately 2.5-Nm.%

To realize the counterbalance design, we used the model as a starting point (Fig.~\ref{fig:comstudy}, blue line), and manually tuned the final counterbalance configuration during construction. The battery bay structure (made from steel) provides both a stiff structure to support the battery and contributes about half of the needed 6 kg counter-mass. Steel plates and bars underneath the battery (Fig.~\ref{fig:torsocad}, left) allow for high-resolution calibration of the counter mass. Proper calibration shows gauge values below 3.0 Nm, a torque achievable by our lower-cost (i.e., less than US \$100), low-profile actuator\footnote{This actuator---a window motor---is also quiet, especially when compared to small but high speed motors with larger gearing.}. %

\subsubsection{Waist Transmission Design}
A non-backdrivable transmission was chosen to minimize the energy required to bow so holding positions requires zero energy. %
An optional locking pin feature allows for the torso motion to be locked for shipping or if waist actuation is not desired.

Friction damper pads on each side of the battery bay (Fig.~\ref{fig:torsocad}, left) add dampening to the waist motion. The pads consist of soft foam and a PTFE sheet fastened to the battery bay, which push against an ABS plastic sheet. This design compensates for gear backlash and compliance in the structure and actuator and greatly simplifies smooth control. The damper increases the torque required to rotate the torso, but this effect was measured empirically to bring the waist motor torque to no more than 3.0 Nm, which met our goal.%

\subsection{Mobile Base Design}
\label{subsubsec:BaseDesign}

\begin{figure}[t]
\centering
\includegraphics[width=3.5in]{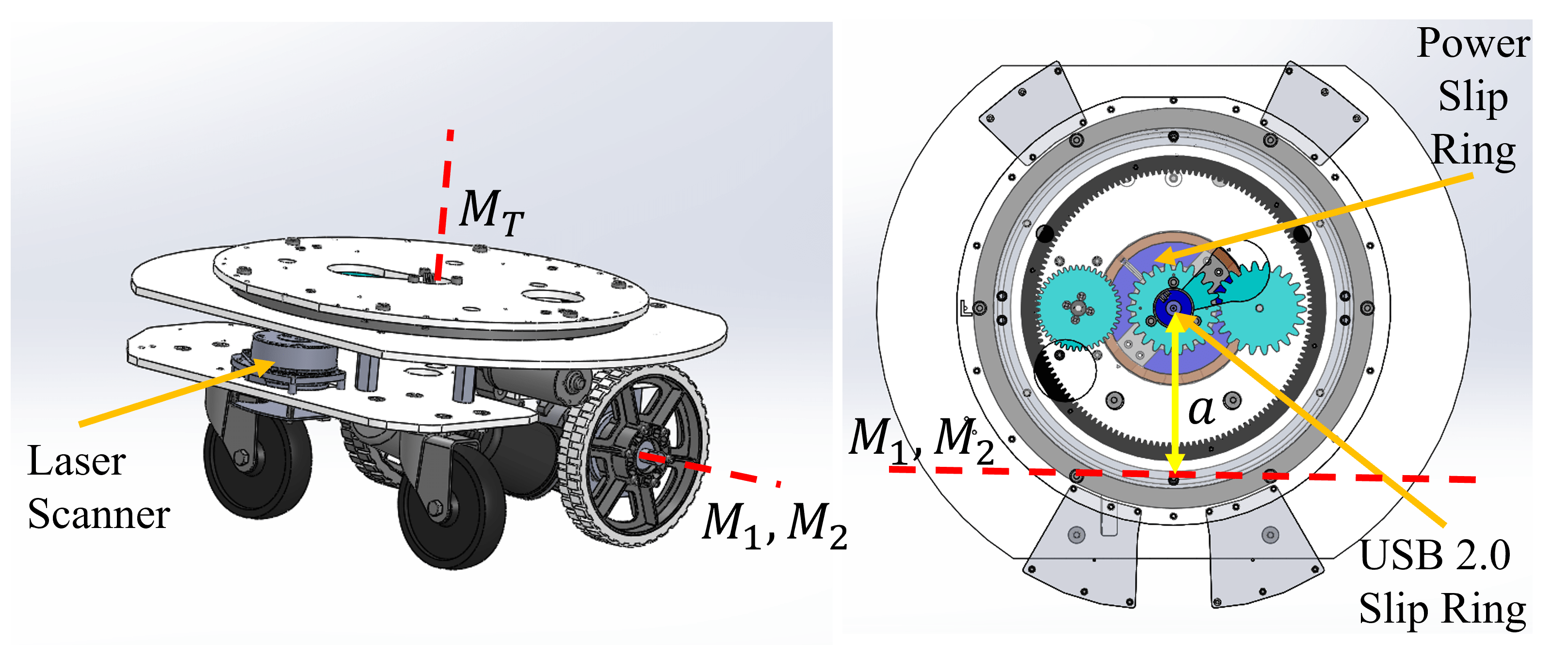}
\caption{Quori base's 3-DoFs produce holonomic movement in the ground plane. The axes driven by the three actuators are highlighted. M$_1$ and M$_2$ are drive motors and are equivalent to a differential drive. M$_T$ is driven by the turret motor. The axis of M$_T$ is distance $a$ from the M$_1$ and M$_2$ axis.}
\label{fig:basecad}
\end{figure}

Quori's holonomic mobile base has three motors (Fig.~\ref{fig:basecad}). %
Two casters serve to support the weight of the robot and increase the support polygon along with two driven wheels. The torso provides electrical power to the base via a slip ring between the turret and differential drive base (Fig.~\ref{fig:basecad}, right). Communication and control between the base and torso electronics occurs via a USB connection through a second concentric slip ring (Fig.~\ref{fig:basecad}, right). Extra space and USB ports are available for a laser scanner or camera in the lower section of the base. %

\begin{figure}[t]
\centering
\includegraphics[width=1.5in]{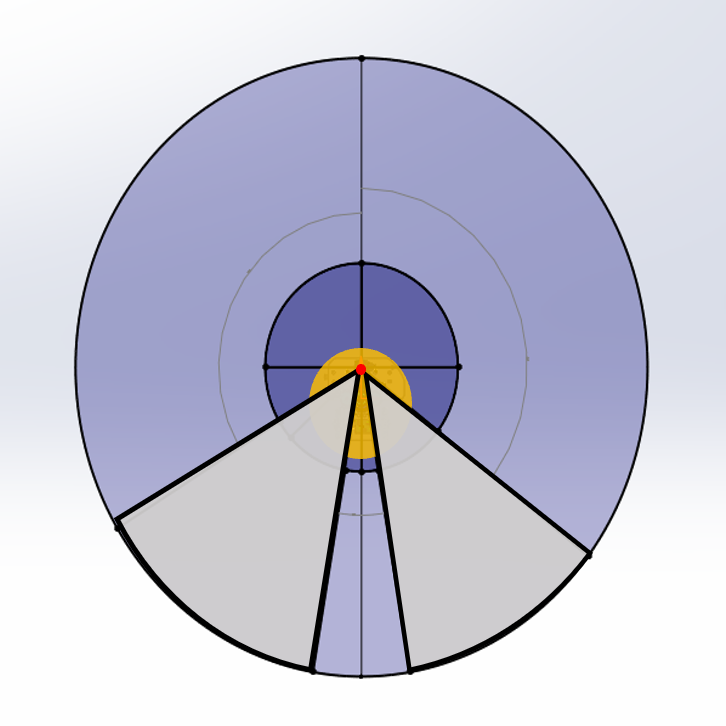}
\setlength{\unitlength}{1in}
\begin{picture}(0,0)
\put(-1.4,0.8){\footnotesize121}
\put(-0.6,0.84){\footnotesize127}
\put(-0.925,0.125){\footnotesize 8}
\put(-0.83,0.125){\footnotesize 8}
\end{picture}
\caption{The laser scanner's FoV. The sensor, marked as a red dot, is offset 100mm to maximize coverage. Sensor blind spots are shaded in gray. The outer circle shows the sensor's 8-meter radius about the robot, marked as a yellow circle.}
\label{fig:ldsfov}
\end{figure}
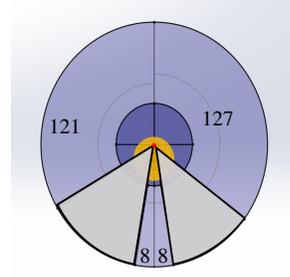

Quori's base measures 48cm in diameter and 20cm in height, and can traverse any indoor floor that complies with the 2010 Americans with Disabilities Act (ADA) Standards for Accessible Design. This includes traversing over 0.635cm bumps (ADA 303.2), 1.27cm floor gaps (ADA 302, 407.4.3), and 1:12 inclines (ADA 405.2). The base has max speeds of 0.6 m/s in a straight line and $\pi \frac{rads}{s}$ rotation of the turret. The design tool presented in \cite{costa2017designing} verified the base parameters will achieve the desired maximum rotational and translational velocities given the motor limits. %
The positioning of the laser ranging sensor near the perimeter of the base maximizes the laser field-of-view (FoV)  (Fig.~\ref{fig:ldsfov}). Finally, the design allows the base to act as a standalone module independent of the upper-body humanoid torso, should the user desire applications with either half alone.

The choice of design for the holonomic base ensures notable cost-reduction over other options; for example, with three motors, our base uses fewer actuators than other designs that require four or more motors \cite{deyle_2010}. Other holonomic designs may involve using an omniwheel or additional motors; however, they often suffer from performance drawbacks, such as vibration or complexity \cite{el2007comparing}. The manufacturing of Quori's base is made more affordable by using laser-cut parts from sheet ABS and commercial off-the-shelf parts for the majority of the components, requiring only two machined parts to mount the motor to the base and the motor shaft to the wheel (Section~\ref{sec:cost}).

\subsection{Power and Electronic Design}
\label{subsec:PowSysDesign}

\subsubsection{Power System}
\label{subsubsec:Power}

Quori's power system was designed to operate untethered with the use of a 12V 40AH battery, which is also used as a counter balance. The battery is a Sealed Lead Acid (SLA) Absorbent Glass Mat (AGM) chemistry battery that is affordable (compared to lithium-based batteries) and stable without losing charge over long periods and has minimal risk of fires or acid spill. %
Shipping is also simplified as the battery only requires the use of a sticker stating ``non-spillable battery'' instead of additional regulations or costs. A potential downside of SLA batteries is the low energy density and high mass; however, we take advantage of this as a counterbalance, as discussed in Section~\ref{subsubsec:TorsoDesign}.

\begin{figure}[t]
\centering
\includegraphics[width=2.5in]{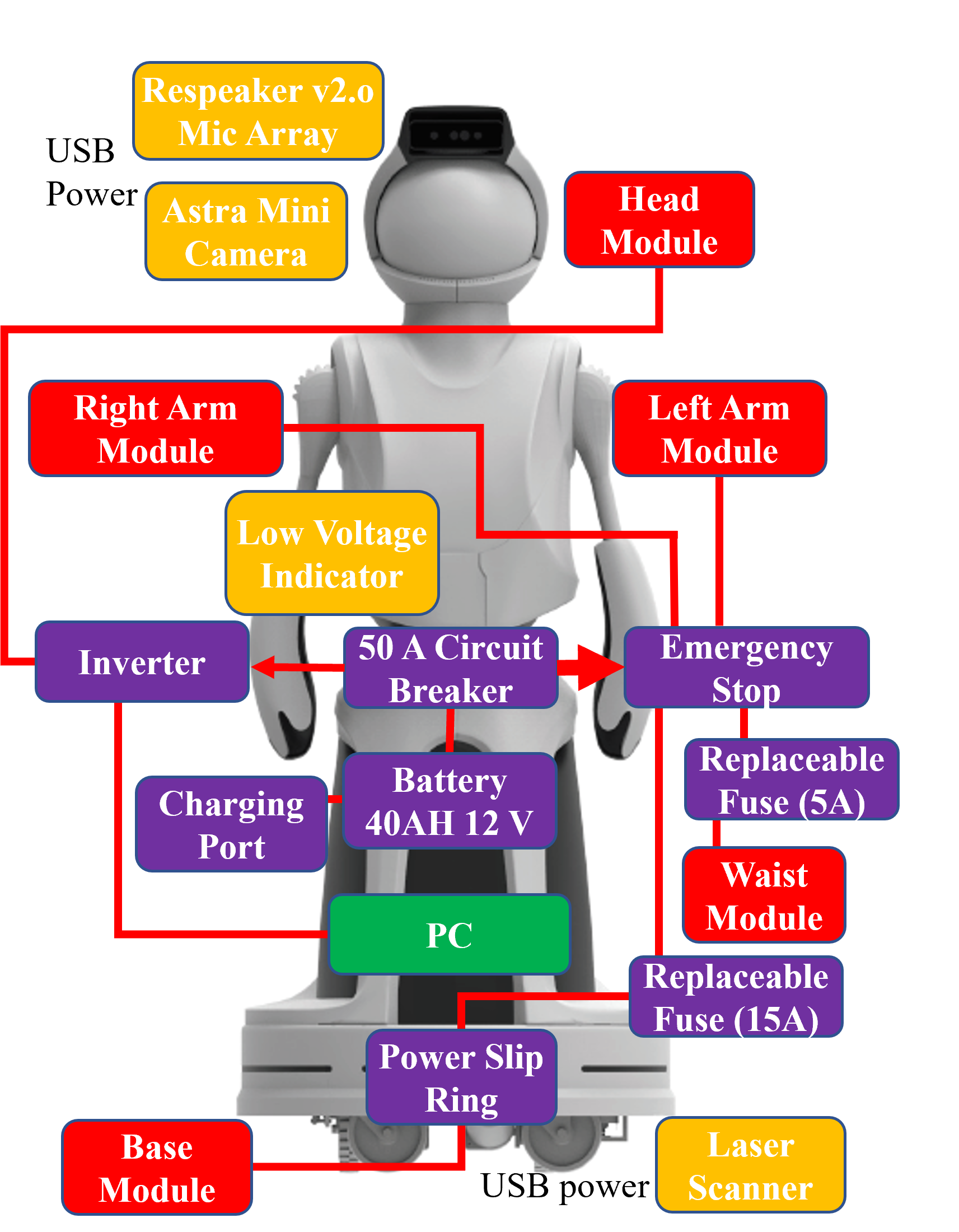}
\caption{The 12V DC circuit for Quori. Motor controllers receive power directly from the battery, while sensors receive power from the computer. The emergency stop controls power to the motors, but allows the computer and projector to remain on. The power charging port is within the battery bay.}
\label{fig:powerOver}
\end{figure}

Most of the robot's subsystems are 12V-based; the only significant voltage switching occurs in a DC-to-AC power inverter, which allows for a main computer (e.g., a laptop) to be used on Quori without requiring the selection or design of an additional DC-to-DC voltage regulator. A simplified diagram of components is presented in Fig.~\ref{fig:powerOver}.
The robot can also run in a tethered mode when not mobile.%

\subsubsection{Electronics Overview}
\label{subsubsec:Ece}
\begin{figure}[t]
\centering
\includegraphics[width=2.5in]{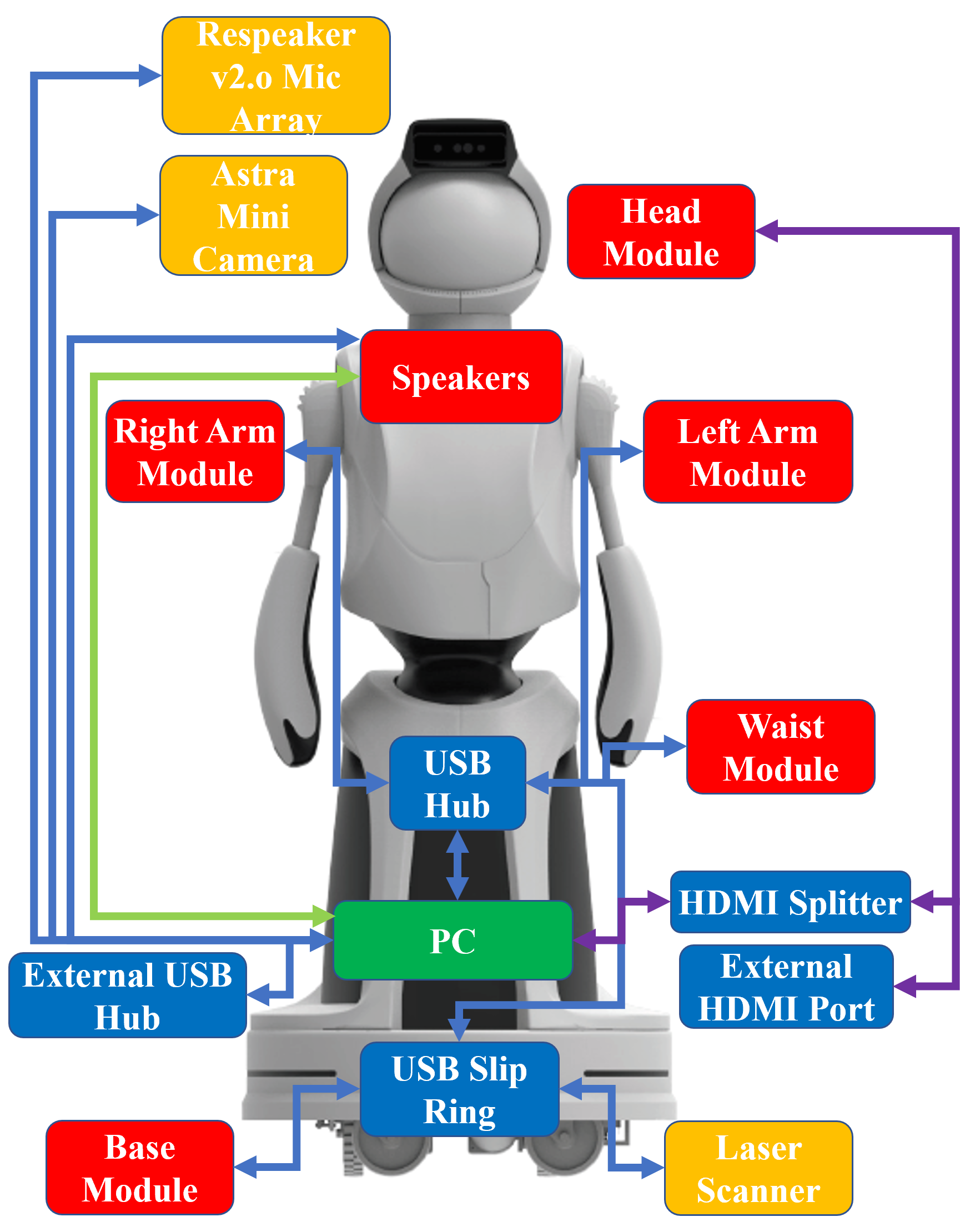}
\caption{Data are transferred via standard methods of USB 2.0 and 3.0 (blue lines), audio jack (green lines) and HDMI (purple lines). Module and sensors are readily modified or replaced with other devices that communicate over USB. A four-port USB hub and a one-port HDMI port are accessible from the back of the robot without removing any components.}
\label{fig:dataOver}
\end{figure}

Each of the sensors and main components connect via standard connectors and communication interfaces for simplicity, modularity, and potential future reconfiguration. The main connection type is USB with all connections using USB 2.0, except for the USB 3.0 RGB+D camera (Quori's default PC has multiple USB 3.0 ports for future upgrades); Fig.~\ref{fig:dataOver} shows the components and connection types. HDMI transmits the head image data, allowing for future modifications. A stereo audio cable from the PC 3.5-mm audio port transmits audio to the chest speakers. A USB and HDMI port are accessible from the back panel of the robot for programming and debugging (Fig.~\ref{fig:3dprinting}, left). 

\subsubsection{Sensors for Social Interaction}
Stereo speakers mount to a shelf on the upper torso behind the chest panel allowing for ample volume (60 dB SPL at 3 meters). Slots in the helmet provide the illusion of sound being produced in the head.

A ReSpeaker 2.0 four-microphone array mounts to the top panel of the helmet for sound localization and speech recognition (Fig.~\ref{fig:headcad}). To test the sensor placement effectiveness for speech recognition, we performed a word error rate (WER) experiment using ten prerecorded English phrases produced from a hardware speaker at three distances (0.1m, 0.5m, and 1.0m), %
yielding an average WER below 13\%. Additional or replacement microphones can be mounted inside the helmet.

\begin{figure}[t]
\centering
\includegraphics[width=3.5in]{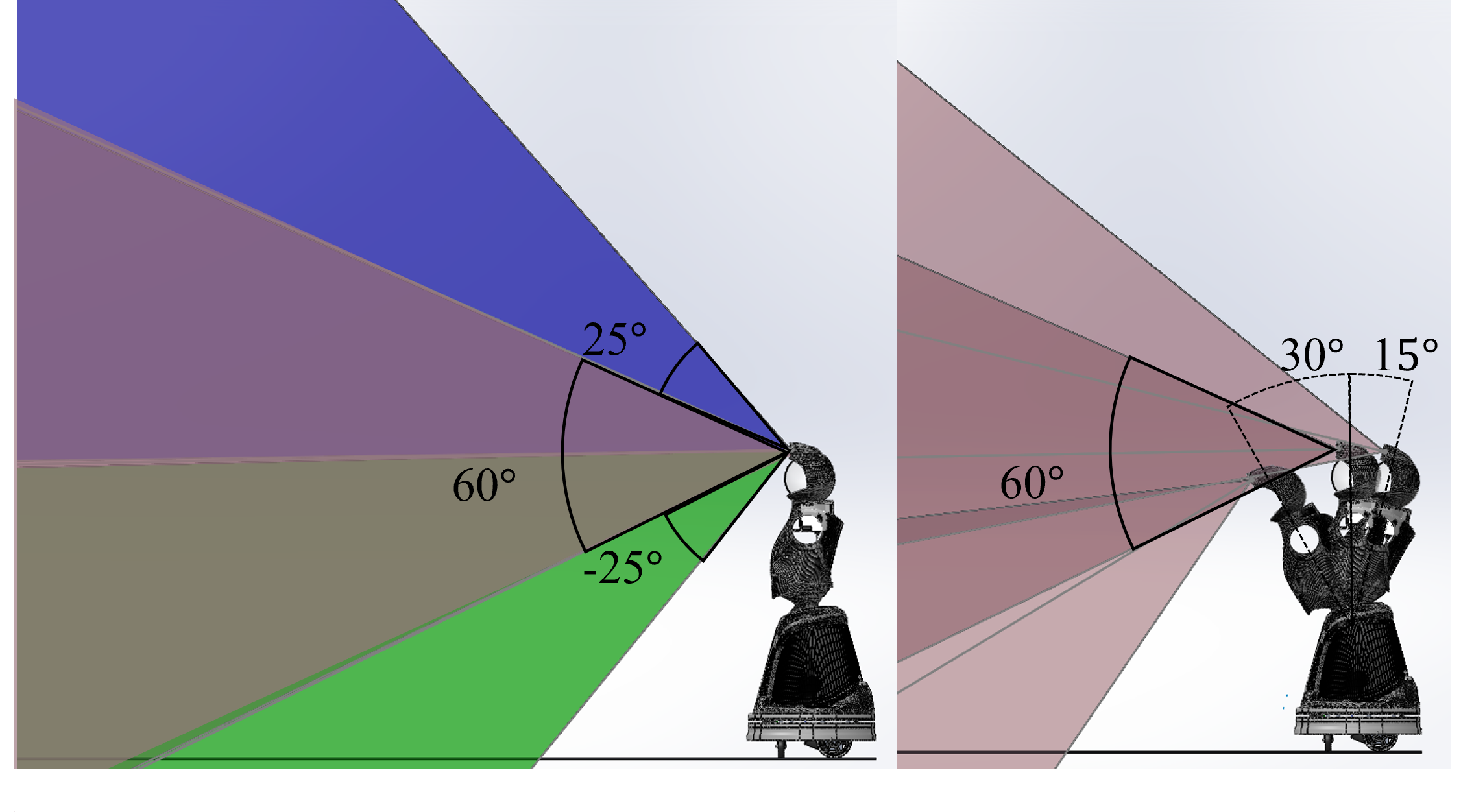}
\caption{Quori's RGB+D FoV can be positioned by two means, manually and by rotation of the torso while the robot is moving. Left: FoV of the camera when positioned manually to neutral(red), $25^{\circ}$ maximum angle up (blue) and down (green). Right: Discrete sweep of the camera FoV at neutral position for three torso positions, $30^{\circ}$ forward, neutral, and $15^{\circ}$ backward.%
}
\label{fig:camerafov}
\end{figure}

An RGB+D camera mounts atop the robot's head to the sensor plate and fits inside the helmet (Fig.~\ref{fig:headcad}). The position of the camera gives the robot a $60^{\circ}$ H x $49.5^{\circ}$V FoV that follows the robot's gaze direction. %
The camera is also plainly visible, which helps to set reasonable social expectations for what is in its FoV (Fig.~\ref{fig:camerafov}); this FoV can be manually adjusted to $\pm 25^{\circ}$ (Fig.~\ref{fig:camerafov}, right). The current camera is an Orbbec Astra Mini\footnote{\url{http://shop.orbbec3d.com/Astra-Mini_p_40.html}} in a DuriPOD Case\footnote{\url{http://shop.orbbec3d.com/DuriPOD_p_47.html}}, which is 120mm x 37.5mm x 32.5mm, but can be easily replaced.

\begin{figure}[t]
\centering
\includegraphics[width=3.5in]{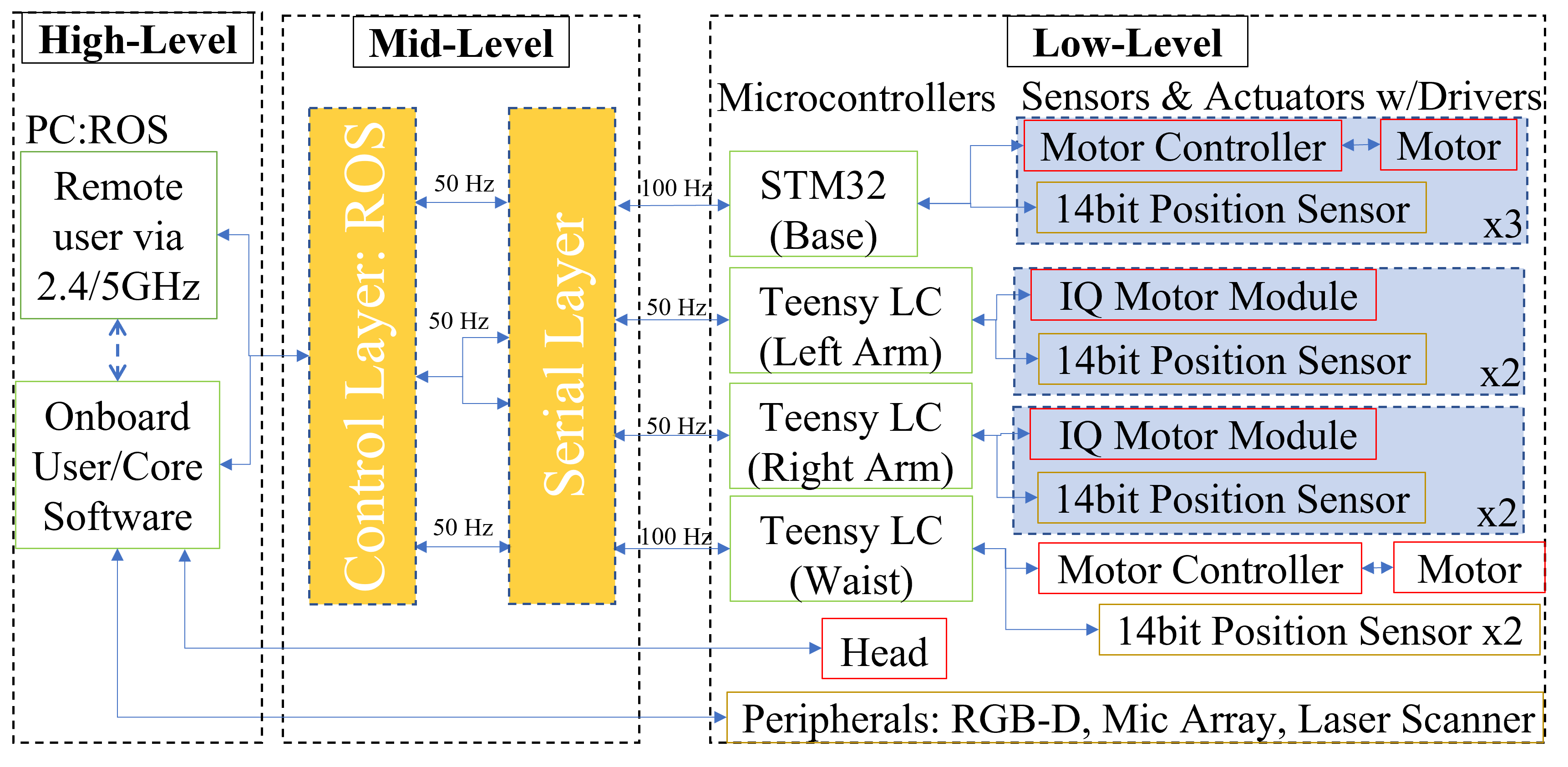}
\caption{Quori robot software system showing basic functionality using ROS for mid-level control of modules. PC usage is in parallel with the microcontrollers that control motor position, speed, measurements, safety, etc.
}
\label{fig:controlSoftwareOver}
\end{figure}

\section{Software Architecture}
\label{sec:Software}
Quori has two main software categories: (1) low-to-mid-level,  including core control of each module (actuation and sensing); and (2) high-level social interaction software (animation and dialog tools), %
as shown in Fig.~\ref{fig:controlSoftwareOver}.

\newpage
\subsection{Low- and Mid-Level Software and Networking} %
\label{subsec:softLowlevel}

The low- and mid-level software, written for ROS \cite{quigley2009ros}, handles low-level actuator voltage commands, communication between microcontrollers and the main PC, and basic control and safety features. Low-level control is handled by microcontrollers, and middle-level control is handled by the onboard PC. The microcontrollers run low-level control independent of the PC. This means safety features (e.g., timers and position limits) are not affected by potential software errors or PC issues. The head module runs at the high level and is \textit{not} discussed in this section. Commands can be sent to the robot and the status can be monitored wirelessly at 2.4 or 5GHz. Quori’s low- and mid-level software capabilities adhere to the ROS developer’s guidelines\footnote{http://wiki.ros.org/DevelopersGuide} to provide an idiomatic experience for the HRI research community.

\subsection{High-Level Social Robotics Software}
\label{subsec:highlevel}
In collaboration with Semio\footnote{https://semio.ai}, we integrated high-level software packages that provide a set of socially interactive behavior APIs and developer tools that can be used by HRI researchers in a platform-agnostic way, analogous to those already in use in commercial products. This software aims to facilitate exploration of advanced topics in HRI without the technical burden of developing and maintaining social behavior primitives on an institutional or individual level. In particular, we provide software packages that enable verbal HRI (via speech recognition and generation), as well as nonverbal HRI (via pointing gesture recognition, and attention recognition and generation); in addition, we provide intuitive software tools to enable HRI research teams to rapidly create and deploy multimodal conversational content on Quori.

\begin{figure}[t]
\centering
\includegraphics[width=3.5in]{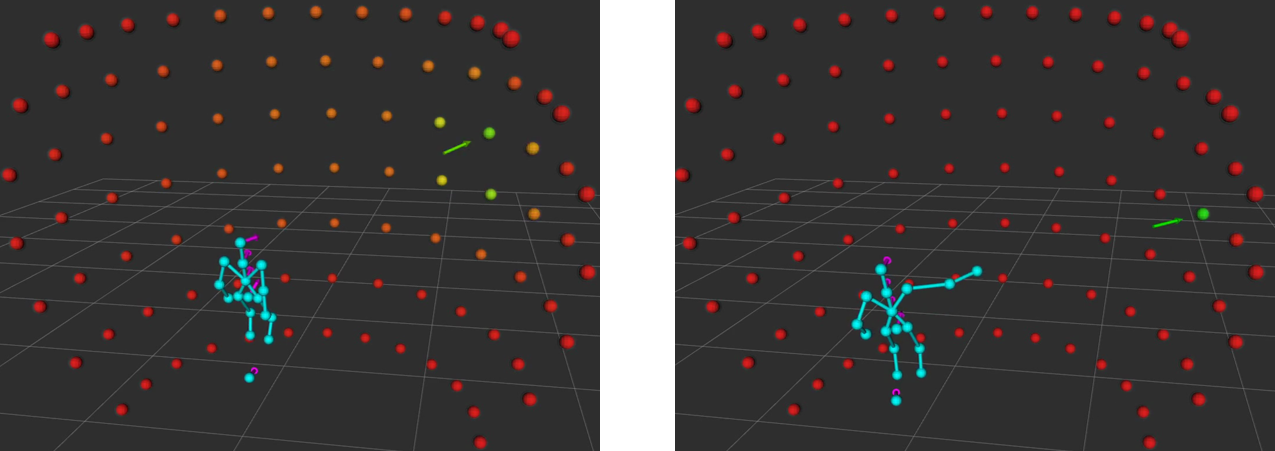}
\caption{Quori's visual attention (left) and pointing gesture recognition (right).}
\label{fig:AttentionDeixis}
\end{figure}

\subsubsection{Speech Generation and Recognition}
\label{subsubsec:speech}
Speech generation and speech recognition were the most desired software modules by HRI researchers in our first survey. While speech generation and recognition are not novel technologies, they are essential in much of social HRI and are therefore key for a standardized platform for HRI research.

\subsubsection{Attention Generation and Recognition}
\label{subsubsec:attention}
The \textit{attention generation} software module uses a mixture of three models of human visual attention to produce robot eye gaze behaviors for face-to-face HRI. These models include (1) a neurobiological model of the human visual attention system \cite{itti2004}, (2) a conversational gaze model for multi-party interactions \cite{mutlu2012}, and (3) a functional gaze model governed by the need for the robot to track body features of the human user to conduct a successful interaction \cite{mead2016}.

The \textit{attention recognition} software module (Fig.~\ref{fig:AttentionDeixis}) provides a probabilistic estimate of targets visually attended to by a human user based on a toe-to-head anatomically correct estimate of human attention (i.e., from human body range of motion to the distribution of photoreceptors in the human eye); distant eye gaze tracking might be pursued in future implementations of this module. The mean error in head pose estimation was determined to be 2.8$^{\circ}$--6.0$^{\circ}$, depending on the data set and environmental conditions. Our solution reduces the overall image search space for faces by 99\% and subsequent attention recognition achieves a frame-rate of 100+ frames per second, which was a limitation of the camera hardware being used (i.e., the highest framerate camera we had available). Furthermore, our software file size is compact, only 110 MB (or up to 160 MB with examples).

\subsubsection{Pointing Gesture Recognition}
\label{subsubsec:pointing}
The static referential \textit{pointing gesture recognition} software module considers the human kinematics of reaching and visual servoing (the relationship between arm kinematics and ``simulated perception'' of visual attention toward an object). The implemented solution yields a mean error in pointing target estimation of 0.17m; it is worth noting that this error is a function of the error rate in human joint tracking from the sensor suite used during testing and not a result of the algorithm.

\subsubsection{Animation and Dialog Tools}
\label{subsubsec:tools}

\begin{figure}[t]
\centering
\includegraphics[width=3.5in]{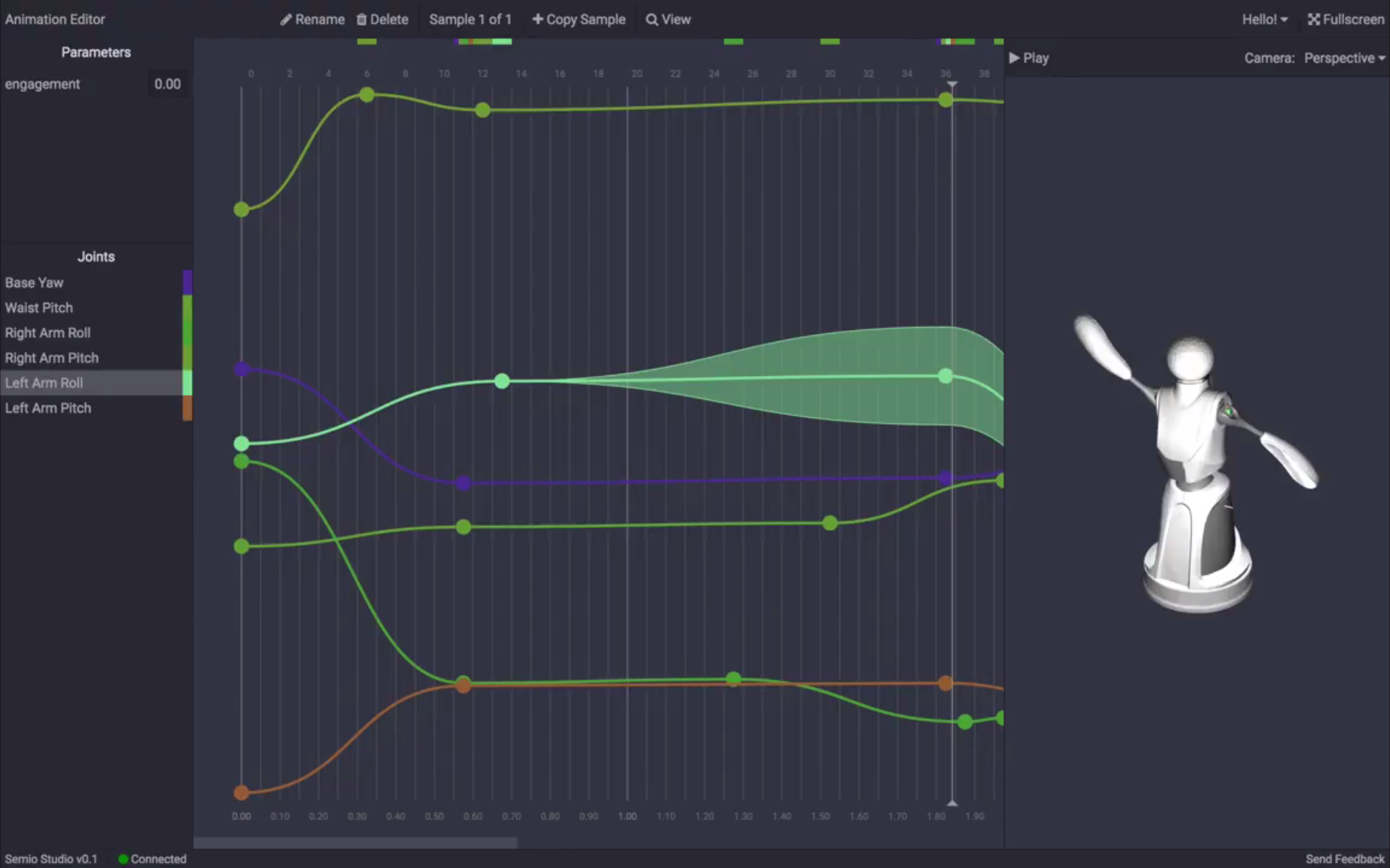}
\caption{Quori's robot keyframe animation tool.}
\label{fig:AnimationEditor}
\end{figure}%

During the community-driven design process, we learned that HRI research groups are often composed of both programmers and non-programmers. Teams interested in using robots for HRI experiments should be able to animate socially interactive robot behaviors and synchronize those behaviors with speech without needing to program, thereby lowering the development barrier for non-programmers. SoftBank Robotics offers similar software tools in their Choregraphe Suite\footnote{https://developer.softbankrobotics.com/nao6/naoqi-developer-guide/choregraphe-suite} and Animation\footnote{https://developer.softbankrobotics.com/pepper-qisdk/tools/animation-editor}/Chat\footnote{https://developer.softbankrobotics.com/pepper-qisdk/tools/chat-editor} Editors; however, those tools are specific to their NAO and Pepper robot platforms, and are not extensible to other robots. To support the Quori community, we developed two web-based tools\footnote{Built using React, Three.js, Node.js, and Typescript.}---one for keyframe animation of expressive robot movement (similar to Maya\footnote{https://www.autodesk.com/products/maya} or Blender\footnote{https://www.blender.org}; Fig.~\ref{fig:AnimationEditor}) and one for authoring human-robot dialog (similar to Voiceflow\footnote{https://www.voiceflow.com} or Dialogflow\footnote{https://dialogflow.com})---for both programmer and non-programmer HRI researchers, to enable them to rapidly create and deploy socially interactive robot applications that rely on human-robot speech and body language. These tools operate within a web browser and do not require content creators to be familiar with or use Linux or ROS, lowering the barrier to entry for socially interactive robot application development. These tools are cross-platform, allowing the resulting applications to be integrated with, deployed to, and executed on a very broad range of platforms in addition to Quori.
These tools and open-source wrappers around their APIs will be open-sourced for non-commercial applications via the web portal on the Quori website. All created content will be available for public use, viewing, copying, and modifying (similar to Wikipedia\footnote{https://www.wikipedia.org}, but for conversational content) to support  replicability in research.

\section{Testing and Robustness}
\label{sec:testing}
Simplicity of the mechanical design and transmissions are key to prevent failures and reduce low-level testing needs. To ensure robustness, we tested the basic function of the robot in stages, as individual modules and also as a fully integrated system in a laboratory setting.
We examined performance metrics for each module, as well as some life-cycle tests. The culminating platform life-cycle test was a deployment in a public setting running typically for 14 hours a day, 7 days a week, for 6 months as part of an exhibit at the Philadelphia Museum of Art.

\subsection{Module Testing and Performance}
\label{subsec:performance}
Table~\ref{table:test} lists the relevant specifications that are verified for each module before a robot is shipped. One of Quori's unique mechanical designs is its arm transmission, which features a friction wheel pair that acts as a clutch and a speed reduction (Fig.~\ref{fig:armcad}). This part required dedicated testing and a redesign. We describe the process to inform future use of this design.

The material choices for the friction pair were originally MDF and urethane following the design in \cite{whitney2014passively}; however, after life cycle testing, the longevity of the friction pair proved too short and the urethane roller failed via wear, being unable to transfer sufficient torque (less than $60\%$ of its original potential). A urethane roller and aluminum wheel were the final materials chosen; this combination reduced wear on the urethane roller and maintained sufficient torque transfer ability after 70 hours of tested motion.

\begin{table}[t]
\caption{Quori's hardware overview.  The DoF column shows actuation and projection (head module) capabilities.} %
\label{table:test}
\begin{tabular}{|l|r|l|l|r|}
   \hline
            & DoFs                   & Joint Limit                                                              & Max Speed                                                                  & Mass (kg) \\
            \hline
Base        & 3                                         & continuous                                                               & \begin{tabular}[c]{@{}l@{}} $\pi~rads/s$\\ 0.6 m/s fwd\end{tabular} & 9.8                           \\
   \hline
Arm         & 2                                         & \begin{tabular}[c]{@{}l@{}}continuous\\ $\pm$ 70$^{\circ}$\end{tabular} &  1.2 $rads/s$                                        & 2.1                           \\
   \hline
Waist/Torso & 1                                         & +0.35$^{\circ}$,-0.17$^{\circ}$                                                                   & 1 $rads/s$                                                     & 29.5                          \\
   \hline
Head        & \begin{tabular}[c]{@{}r@{}}$inf.$\\ $\sim90$k px \end{tabular}& fixed                                                                    & -                                                                          & 2.0                           \\
   \hline
System      & 8                                         & -                                                                        & -                                                                          & 45.5                         \\
   \hline
\end{tabular}

\end{table}

\subsection{Deployment at Philadelphia Museum of Art}
\label{subsec:pma}
\begin{figure}[t]
\centering
\includegraphics[width=3.2in]{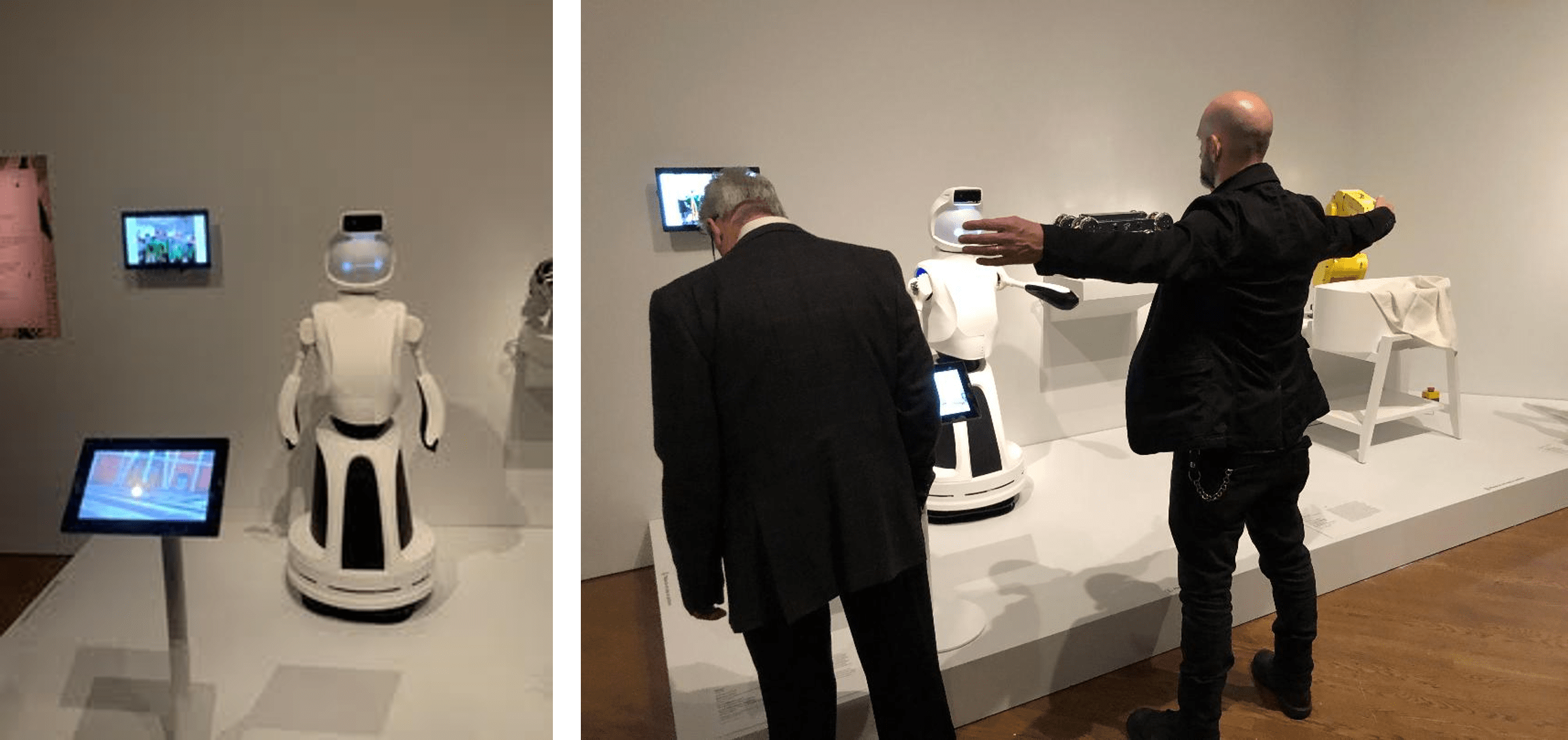}
\caption{Quori was installed at the Designs for Different Futures exhibit at the Philadelphia Museum of Art  from October 2019 to March 2020. The robot autonomously interacted with visitors and reacted with facial expressions, arm gestures, bowing, and tracking visitors by rotating the base.%
}
\label{fig:pma}
\end{figure}

Quori was installed at the ``Designs for Different Futures'' exhibit at the Philadelphia Museum of Art from October 2019 to March 2020, and attracted over 183,000 visitors (Fig.~\ref{fig:pma}). The curators chose Quori as an example of ``a robot of the future''. Quori autonomously interacted with visitors and reacted with animated facial expressions, arm gestures, torso bowing, and tracking guests by rotating the base while staying fixed on the platform. An external monitor showed visitors a sample of what Quori could see along with a kinematic overlay of identified humans' limbs in its FoV. This is the first application of Quori as a socially interactive robot platform. More applications and evaluations will follow as the Quori platforms are distributed to researchers.

Highlights of Quori's reactions included waving ``hello'' to visitors who entered its FoV, dancing to gain attention, and bowing to greet visitors. The robot attempted to stay engaged with the closest visitor by tracking them with the base turret actuator. Quori then attempted to mirror the visitor's arm movements (Fig.~\ref{fig:pma}, right). If no one was in Quori's FoV, it returned to an ``sleep'' position by rotating its torso to a center position and leaning over with its arms hanging and the face switching to a loading animation. The interaction with visitors was based on a finite state machine with action cool-downs and was designed to feel spontaneous without being repetitive. %

Quori's hardware performed well overall. Quori was active 7 days a week, from 8am--10pm. A power supply and power strip powered the robot, and its system turned itself on and off each day to reduce strain. Museum staff reported any issues daily, which mostly consisted of synchronization with the external monitor. Specifically, the museum required a total of 11 visits to address system issues; of those issues, 6 were related to motor drive code failure, 2 to other hardware issues, and 3 were software timing issues due to syncing multiple computers and monitors. The motor driver firmware was updated to prevent further failures. A weekly maintenance visit included those repairs as well as inspections of the whole system. 

These visits led to a few design changes that allow for quicker and easier access to the robot, including: (1) replacing key fasteners on the torso panels with thumbscrews instead of socket heads, and (2) selecting a thinner material for the base service panel, which allowed for simpler removal and mounting. Quori's hardware and software were able to perform well over a long-term installation in a public setting, and the experience enabled us to identify weaknesses and improve Quori's hardware and software subsystems.%

\section{Conclusion and Future Work}
\label{sec:conclusion}
This paper presents Quori, an affordable socially interactive robot platform comprised of an upper-body humanoid with a rear-projection head, a one-DoF waist, and two gesturing arms, along with a holonomic mobile base. We describe the features and utility of the four modules and identify the decisions resulting in an affordable design. The modules were designed to produce an aesthetically pleasing whole that meets the requirements identified by the HRI community in surveys and workshops. %

The ten Quori robots have been awarded to a diverse group of researchers across the United States with many multidisciplinary and cross-laboratory researchers. The Quori website\footnote{\url{http://www.quori.org/community#research-groups}} includes information about the awardees, their research, and updates. Additional dissemination is planned through a Quori simulation using the Gazebo\footnote{http://gazebosim.org} 3D robot simulator.

Further evaluation and assessment of the Quori platform will be possible after those ten research groups have received the robots and have had the time to develop new capabilities and perform user studies. In future work, we will perform a full-scale assessment of Quori's effectiveness in meeting the needs of the HRI research community.

\section{Acknowledgements}
This work is supported by the National Science Foundation under Grant No. CNS-1513275 and CNS-1513108.


\bibliographystyle{IEEEtran}
\bibliography{IEEEabrv,references.bib}

\begin{IEEEbiography}[{\includegraphics[width=1in,height=1.25in,clip,keepaspectratio]{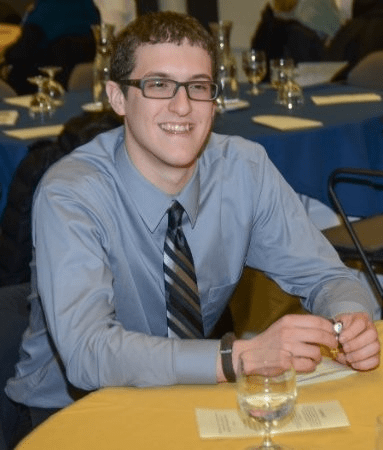}}]
{Andrew Specian}
is a Ph.D. student at the University of Pennsylvania, PA, USA in the
  ModLab, a part of the GRASP Laboratory. His research focuses on novel manipulator design and planning for robot capabilities. His research interests include designing robots for Search and Rescue and Human Robot Interaction.
\end{IEEEbiography}

\begin{IEEEbiography}[{\includegraphics[width=1in,height=1.25in,clip,keepaspectratio]{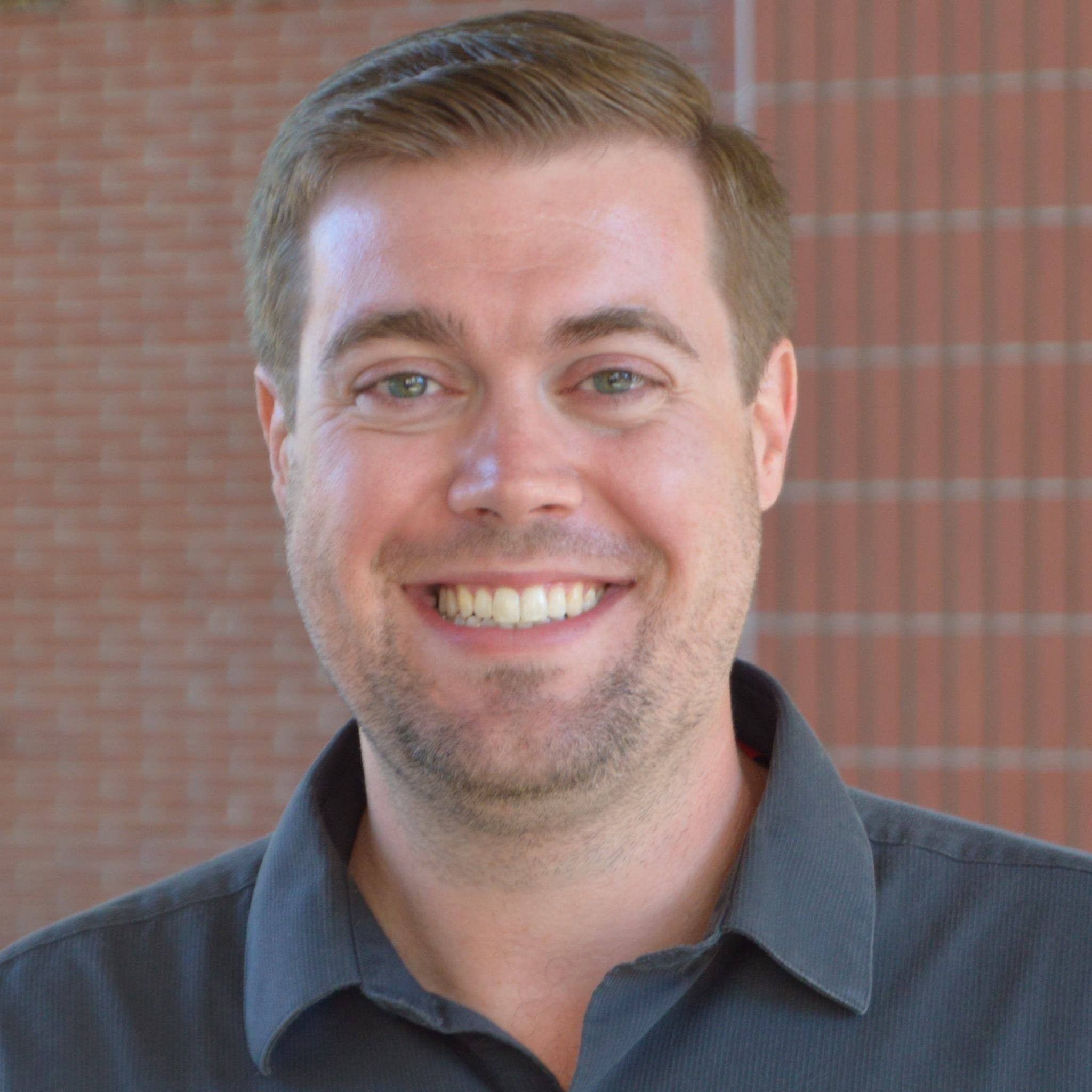}}]
{Ross Mead}
is the Founder and CEO of Semio AI, Inc. (Semio). Ross received his PhD and MS in Computer Science from the University of Southern California in 2015, and his BS in Computer Science from Southern Illinois University Edwardsville in 2007.  Ross’s dissertation work focused on socially assistive robots, specifically, on the principled design and computational modeling of fundamental social behaviors (such as speech, gestures, eye gaze, social spacing, etc.) that serve as building blocks for automated recognition and control in face-to-face human-robot interactions. His research provides the foundations upon which Semio software is being built.
\end{IEEEbiography}

\begin{IEEEbiography}[{\includegraphics[width=1in,height=1.25in,clip,keepaspectratio]{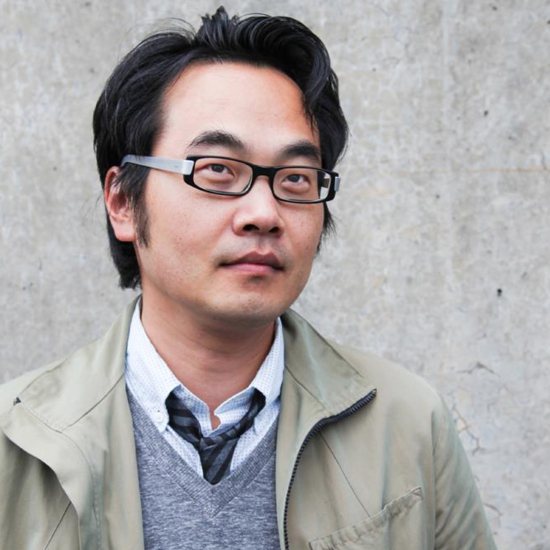}}]
{Simon Kim}
is a licensed architect and researcher in applied sciences within the disciplines of architecture and urbanism. Director of the Immersive Kinematics Lab and Principal of Ibañez Kim, his projects are funded by the National Science Foundation (NSF), Pew Center for Arts and Humanities, Social Sciences and Humanities Research Council (SSHRC), Canadian Heritage Foundation. He is also supported by residencies and fellowships at Autodesk, RAIR Philadelphia, MIT, and the Seoul Biennale. His graduate courses have partnerships with Seoul National University, Opera Philadelphia, Tyler School of Art, with sponsorship from Heerim Architects and Planners. Simon has published and presented his work globally at conferences such as IEEE, AI-HRI, ACADIA, eCAADe, ACSA, SmartGeometry, and for clients such as the MIT Museum, the Philadelphia Museum of Art, Harvard University, the Smithsonian, MoMA, Slought Foundation.
\end{IEEEbiography}

\begin{IEEEbiography}[{\includegraphics[width=1in,height=1.25in,clip,keepaspectratio]{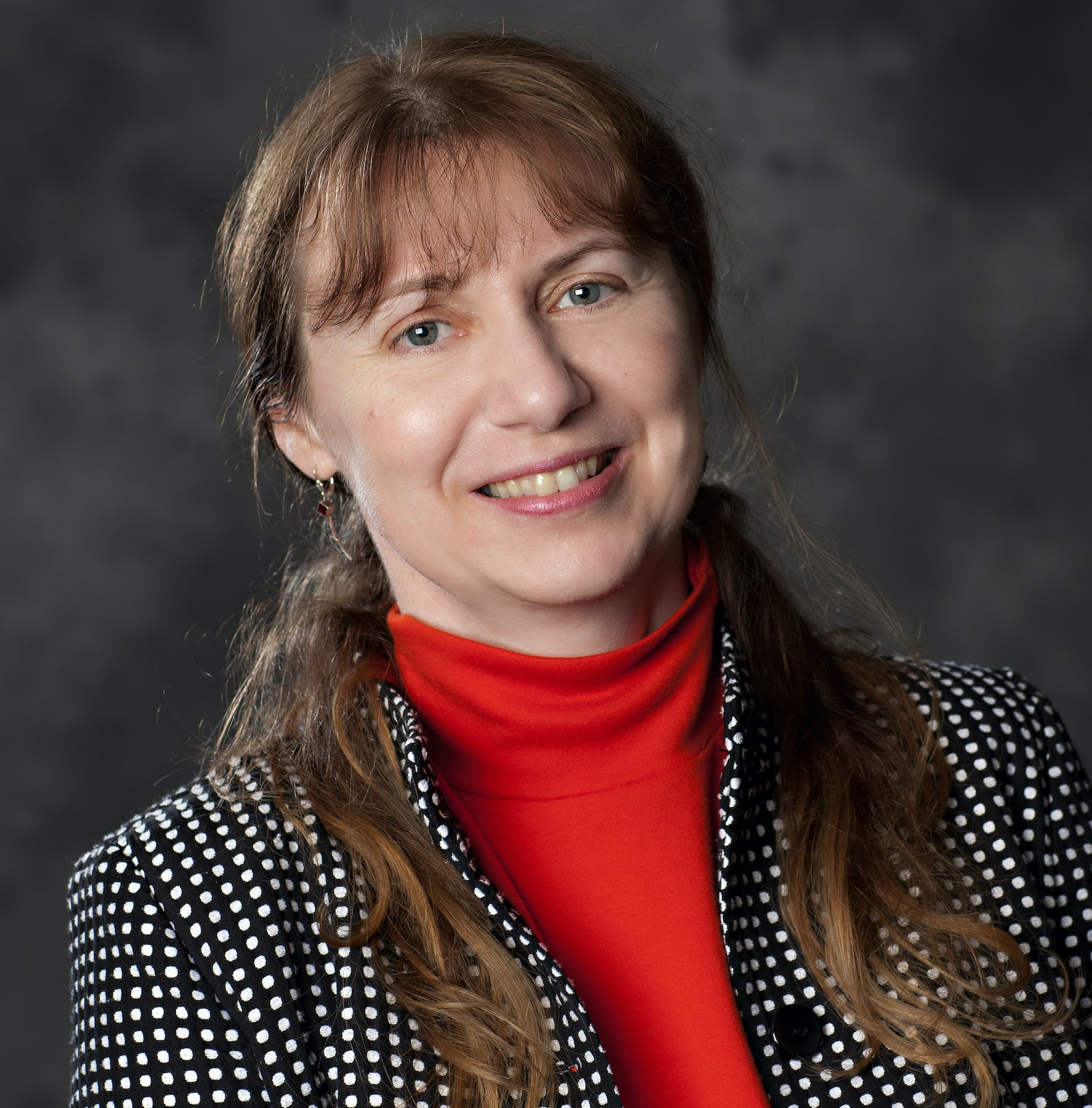}}]
{Maja Matari\'c}
is Chan Soon-Shiong Professor of Computer Science, Neuroscience, and Pediatrics at USC, founding director of the Robotics and Autonomous Systems Center, and lead of the Viterbi K-12 STEM Center. Her PhD and MS are from MIT, and her BS is from Kansas University. She is Fellow of AAAS, IEEE, AAAI, and ACM, recipient of the Presidential Award for Excellence in Science, Mathematics \& Engineering Mentoring, Anita Borg Institute Women of Vision for Innovation, NSF Career, MIT TR35 Innovation, and IEEE RAS Early Career Awards, is active in K-12 outreach, and authored ``The Robotics Prime'' with MIT Press. A pioneer of the filed of socially assistive robotics, her research is developing human-robot interaction for convalescence, rehabilitation, training, and education for children with autism spectrum disorders, stroke and traumatic brain injury survivors, and individuals with Alzheimer's Disease. She is also co-founder of Embodied, Inc.
\end{IEEEbiography}

\begin{IEEEbiography}[{\includegraphics[width=1in,height=1.25in,clip,keepaspectratio]{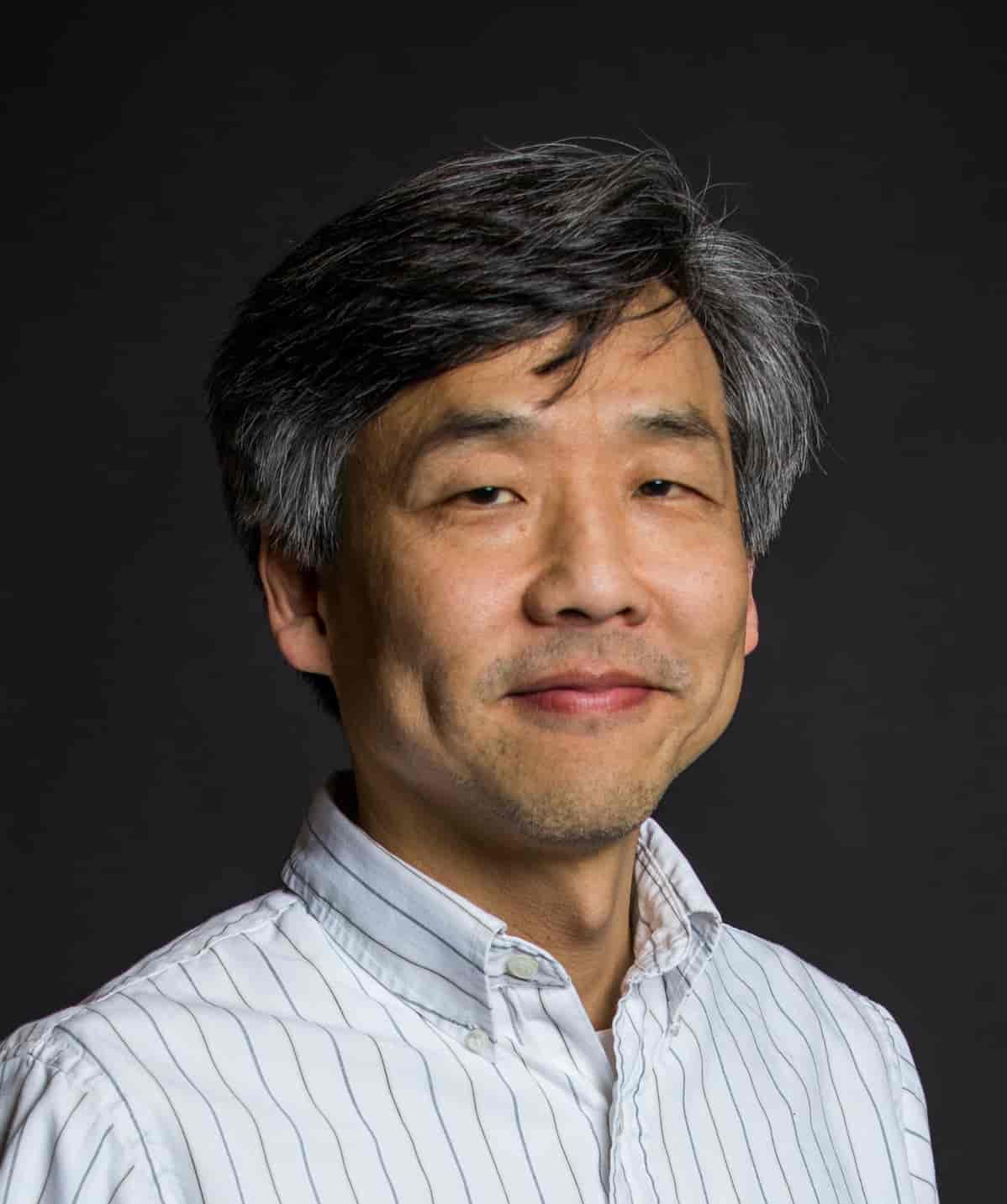}}]
{Mark Yim}
is the Asa Whitney professor of Mechanical Engineering who joined the University of Pennsylvania in 2004. He is the director of the GRASP Lab, the oldest robotics research lab in the country established 1980.  His research group designs and builds a variety of electromechanical hardware. Demonstrations range from a humanoid robot on display at the Philadelphia Museum of Art to transforming robots that can change their shape to the smallest self-powered flying robot in the world. Honors include the Lindback Award for Distinguished Teaching (UPenn's highest teaching honor); induction to MIT's TR100 in 1999; induction to the National Academy of Inventors in 2018.  He has over 250 publications and patents. He has started two companies, one in robotics and one medical device company making a steerable needle.

\end{IEEEbiography}

 
\onecolumn
 
\appendices
\label{Appendix}
 
\section{Survey \#1}
\label{Appendix:Survey1}
 
The first web survey was sent to 37 HRI researchers in Fall 2014. Result ratios below 37 represent the prompt includes skipped response(s). Tables~\ref{table:survey1-app-act}-\ref{table:survey1-dem} present the prompts, responses, and results from the first survey that are related to Quori's design. Table~\ref{table:survey1-app-act} presents the survey data for robot appearance and actuation considerations in support of Section~\ref{subsubsec:surveyact}. Table~\ref{table:survey1-sensing} presents the survey data for robot sensing and behavior considerations in support of Section~\ref{subsubsec:surveysensing}. Table~\ref{table:survey1-cost} presents the survey data for robot cost considerations in support of Section~\ref{subsubsec:surveycost}. Table~\ref{table:survey1-dem} presents demographic data of Survey~\#1.
\begin{table}[ht]
\centering
\caption{Survey \#1: Robot Appearance and Actuation Considerations}
\label{table:survey1-app-act}
\begin{tabular}{|m{3in}cc|}
\hline
Prompt                                                                                                                               & Responses              & Result         \\ \hline
\multirow{6}{3in}{(1) Please select a preferred appearance of the HRI robot platform.}                                                     & Mechanical             & 33\% (12/36)   \\
                                                                                                                                     & Cartoon                & 31\% (11/36)   \\
                                                                                                                                     & Other                  & 14\% (5/36)    \\
                                                                                                                                     & Creature               & 11\% (4/36)    \\
                                                                                                                                     & Human                  & 6\% (2/36)     \\
                                                                                                                                     & No Preference          & 6\% (2/36)     \\ \hline
\multirow{5}{3in}{(2) Please select a preferred outer covering of the   HRI robot platform.}                                               & Hard                   & 50\% (18/36)   \\
                                                                                                                                     & Soft                   & 28\% (10/36)   \\
                                                                                                                                     & No Preference          & 11\% (4/36)    \\
                                                                                                                                     & Fabric                 & 6\% (2/36)     \\
                                                                                                                                     & Fuzzy                  & 6\% (2/36)     \\ \hline
\multirow{3}{3in}{(3) Please select height preferences for the HRI robot platform. Reported in meters.}                                            & Minimum                & 0.71$\pm$0.36 m \\ 
                                                                                                                                     & Preferred               & 1.14$\pm$0.35 m \\
                                                                                                                                     & Maximum                & 1.48$\pm$0.33 m \\ \hline
\multirow{5}{3in}{(4) Please select a preferred gender   of the HRI robot platform.}                                                       & Customizable           & 49\% (18/37)   \\
                                                                                                                                     & Neutral                & 41\% (15/37)   \\
                                                                                                                                     & No Preference          & 5\% (2/37)     \\
                                                                                                                                     & Female                 & 3\% (1/37)     \\
                                                                                                                                     & Male                   & 3\% (1/37)     \\ \hline
\multirow{7}{3in}{(5) Please rank the following   actuation capabilities in order of preference for the HRI robot platform.}               & Head                   & 6.00           \\
                                                                                                                                     & Face                   & 5.00           \\
                                                                                                                                     & Arms                   & 4.58           \\
                                                                                                                                     & Mobile Base            & 4.08           \\
                                                                                                                                     & Trunk/Spine            & 3.28           \\
                                                                                                                                     & Hands                  & 3.08           \\
                                                                                                                                     & Shoulders              & 2.00           \\ \hline
\multirow{7}{3in}{(6) Please rank the following face   actuation capabilities in order of preference for the HRI robot platform.}          & Eyes                   & 6.43           \\
                                                                                                                                     & Eyelids                & 4.91           \\
                                                                                                                                     & Eyebrows               & 4.69           \\
                                                                                                                                     & Lips                   & 4.56           \\
                                                                                                                                     & Jaw                    & 3.82           \\
                                                                                                                                     & Ears                   & 2.79           \\
                                                                                                                                     & Nose                   & 1.41           \\ \hline
\multirow{4}{3in}{(7) Please rank the following head actuation   capabilities in order of preference for the HRI robot platform.}          & Nodding                & 3.59           \\
                                                                                                                                     & Shaking                & 2.82           \\
                                                                                                                                     & Tilting                & 2.45           \\
                                                                                                                                     & Squashing / Stretching & 1.18           \\ \hline
\multirow{4}{3in}{(8) Please rank the following   trunk/spine actuation capabilities in order of preference for the HRI robot   platform.} & Leaning Forward / Back & 3.41           \\
                                                                                                                                     & Leaning Left / Right   & 2.57           \\
                                                                                                                                     & Twisting Left / Right  & 2.42           \\
                                                                                                                                     & Squashing / Stretching & 1.77           \\ \hline
\multirow{3}{3in}{(9) Please rank the following mobile   base actuation capabilities in order of preference for the HRI robot   platform.} & Omni-drive             & 2.64           \\
                                                                                                                                     & Diff-drive             & 2.06           \\
                                                                                                                                     & Legs                   & 1.34           \\ \hline        
\end{tabular}
\end{table}
\begin{table}[ht]
\centering
\caption{Survey \#1: Robot Sensing and Behavioral Considerations}
\label{table:survey1-sensing}
\begin{tabular}{|m{3in}cc|}
\hline
Prompt                                                                                                                                          & Responses                        & Result \\ \hline
\multirow{10}{3in}{(1) Please rank the following   autonomous behavior generation capabilities in order of preference for the   HRI robot platform.}  & Speech \& Dialog                 & 7.63   \\
                                                                                                                                                & Eye Gaze \& Attention            & 7.37   \\
                                                                                                                                                & Turn-taking \&   Back-channeling & 6.06   \\
                                                                                                                                                & Environmental Navigation         & 6.00    \\
                                                                                                                                                & Social Navigation                & 5.67   \\
                                                                                                                                                & Gestures                         & 5.53   \\
                                                                                                                                                & Emotion                          & 4.89   \\
                                                                                                                                                & Prosody                          & 4.89   \\
                                                                                                                                                & Object Manipulation              & 4.51   \\
                                                                                                                                                & Touch                            & 3.49   \\ \hline
\multirow{11}{3in}{(2) Please rank the following   autonomous behavior recognition capabilities in order of preference for the   HRI robot platform.} & Speech                           & 8.75   \\
                                                                                                                                                & Eye Gaze \& Attention            & 7.44   \\
                                                                                                                                                & Person Identification            & 7.06   \\
                                                                                                                                                & Gesture                          & 6.89   \\
                                                                                                                                                & Object Identification            & 6.15   \\
                                                                                                                                                & Mapping \& Localization          & 6.06   \\
                                                                                                                                                & Social Navigation                & 5.89   \\
                                                                                                                                                & Turn-taking \&   Back-channeling & 5.86   \\
                                                                                                                                                & Emotion                          & 5.44   \\
                                                                                                                                                & Touch                            & 3.91   \\
                                                                                                                                                & Prosody                          & 3.42   \\ \hline
\multirow{6}{3in}{(3) Please rank the following sensing   capabilities in order of preference for the HRI robot platform.}                            & RGB+Depth Camera(s)              & 5.53   \\
                                                                                                                                                & Microphones                      & 4.81   \\
                                                                                                                                                & Distance                         & 3.49   \\
                                                                                                                                                & Tactile                          & 2.86   \\
                                                                                                                                                & RGB-only Camera(s)               & 2.32   \\
                                                                                                                                                & Depth-only Camera(s)             & 2.23   \\ \hline
\end{tabular}
\end{table}

\begin{table}[ht]
\centering
\caption{Survey \#1: Robot Cost Considerations}
\label{table:survey1-cost}
\begin{tabular}{|m{3in}cc|}
\hline
Prompt                                                                              & Responses            & Result         \\ \hline
\multirow{9}{3in}{(1) How much would you expect to pay for the HRI   robot platform.}     & \$25K-50K            & 25\% (9/36)    \\
                                                                                    & \$5K-10K             & 22\% (8/36) \\
                                                                                    & \$2.5K-5K            & 19\% (7/36) \\
                                                                                    & \$10K-25K            & 14\% (5/36) \\
                                                                                    & \$100K-250K          & 8\% (3/36)  \\
                                                                                    & \$1K-2.5K            & 6\% (2/36)  \\
                                                                                    & \$50K-100K           & 6\% (2/36)  \\
                                                                                    & \textless{}\$1K      & 0\% (0/36)     \\
                                                                                    & \textgreater{}\$250K & 0\% (0/36)     \\ \hline
\multirow{9}{3in}{(2) How much would you be willing to   pay for the HRI robot platform.} & \$1K-2.5K            & 17\% (6/36)    \\
                                                                                    & \$5K-10K             & 17\% (6/36)    \\
                                                                                    & \$10K-25K            & 17\% (6/36)    \\
                                                                                    & \$25K-50K            & 17\% (6/36)    \\
                                                                                    & \$50K-100K           & 17\% (6/36)    \\
                                                                                    & \$2.5K-5K            & 11\% (4/36)    \\
                                                                                    & \textless{}\$1K      & 6\% (2/36)     \\
                                                                                    & \$100K-250K          & 0\% (0/36)     \\
                                                                                    & \textgreater{}\$250K & 0\% (0/36)     \\ \hline
\end{tabular}
\end{table}

\begin{table}[ht]
\centering
\caption{Survey \#1: Demographics}
\label{table:survey1-dem}
\begin{tabular}{|m{3in}cc|}
\hline
Prompt                                                                                                                & Responses                 & Result       \\ \hline
\multirow{4}{3in}{(1) What is your age (in years)?}                                                                         & 25-34                     & 53\% (18/34) \\
                                                                                                                      & 18-24                     & 18\% (6/34)  \\
                                                                                                                      & 45+?                      & 15\% (5/34)  \\
                                                                                                                      & 35-44                     & 15\% (5/34)  \\ \hline
\multirow{3}{3in}{(2) What is your gender?}                                                                                 & Female                    & 50\% (17/34) \\
                                                                                                                      & Male                      & 47\% (16/34) \\
                                                                                                                      & Did not specify           & 3\% (1/34)   \\ \hline
\multirow{4}{3in}{(3) What is your ethnicity? (Please   select all that apply.)}                                            & White / Caucasian         & 76\% (26/34) \\
                                                                                                                      & Asian or Pacific Islander & 21\% (7/34)  \\
                                                                                                                      & Black or African American & 3\% (1/34)   \\
                                                                                                                      & Prefer not to specify     & 3\% (1/34)   \\ \hline
\multirow{8}{3in}{(4) How much do you know about the   following topics?}                                                   & Human-Robot Interaction   & 6.38         \\
                                                                                                                      & Robotics                  & 6.24         \\
                                                                                                                      & Artificial Intelligence   & 5.85         \\
                                                                                                                      & Psychology                & 4.65         \\
                                                                                                                      & Signal Processing         & 4.59         \\
                                                                                                                      & Animation (2D/3D)         & 3.97         \\
                                                                                                                      & Anatomy                   & 3.18         \\
                                                                                                                      & Anthropology              & 2.91         \\ \hline
\multirow{2}{3in}{(5) Would you be willing to provide a   letter of support for the development of the HRI robot platform?} & No                        & 50\% (15/30) \\
                                                                                                                      & Yes                       & 50\% (15/30) \\ \hline
\end{tabular}
\end{table}

\clearpage

\section{Survey \#2}
\label{Appendix:Survey2}
The second web survey was sent to 50 HRI researchers in Fall 2015. Result ratios below 50 represent the prompt includes skipped response(s). Table~\ref{table:survey2-app-act}-\ref{table:survey2-dem} present the prompts, responses, and results from the second survey that are related to Quori's design. Table~\ref{table:survey2-app-act} presents the survey data for robot appearance and actuation considerations in support of Section~\ref{subsubsec:surveyact}. Table~\ref{table:survey2-sen} presents the survey data for robot sensing and behavior considerations in support of Section~\ref{subsubsec:surveysensing}. Table~\ref{table:survey2-dem} presents demographic data of Survey \#2.

\begin{table}[ht]
\centering
\caption{Survey \#2: Appearance and Actuation Considerations}
\label{table:survey2-app-act}
\begin{tabular}{|m{3in}m{1.5in}c|}
\hline
Prompt                                                                                                                                                                                                                                                                                                   & Responses                                                               & Result          \\ \hline
\multirow{9}{3in}{(1) By default, Quori will have a mechanical appearance (clearly a robot, not a human); however, users may prefer a cartoonish appearance (similar to a 3D-animated character). What features would best express a cartoonish appearance?}                                             & Projected Face   (e.g., 3D-animated vs. “robotic”)                      & 7.67            \\
                                                                                                                                                                                                                                                                                                         & Vocal   Behavior (e.g., how things are said)                            & 7.36            \\
                                                                                                                                                                                                                                                                                                         & Visual   Behavior (e.g., movement is different)                         & 7.33            \\
                                                                                                                                                                                                                                                                                                         & Arms   (e.g., exaggerated plastic arm covers)                           & 5.29            \\
                                                                                                                                                                                                                                                                                                         & Chest   (e.g., exaggerated plastic chest cover)                         & 5.00            \\
                                                                                                                                                                                                                                                                                                         & Hands   (e.g., exaggerated plastic hand covers)                         & 4.82            \\
                                                                                                                                                                                                                                                                                                         & Mobile   Base (e.g., exaggerated plastic base cover)                    & 3.60            \\
                                                                                                                                                                                                                                                                                                         & Other   (please explain)                                                & 2.31            \\
                                                                                                                                                                                                                                                                                                         & Do Not   Support (please explain)                                       & 2.00            \\ \hline
\multirow{4}{3in}{(2) By default, Quori’s upper body will be   approximately 0.6-0.9 meters (2-3 feet) tall and the mobile base will be   approximately 0.3 meters (1 foot) tall. Combined, Quori will be 0.9-1.2   meters (3-4 feet) tall. How would you prefer the HRI robot platform’s height   to change?} & Telescoping Spine (e.g., the robot can translate up and down a pole.) & 45\% (17/38)  \\
                                                                                                                                                                                                                                                                                                         & Flexible   Spine (e.g., the robot can lean forward and backward)        & 26\% (10/38)  \\
                                                                                                                                                                                                                                                                                                         & No   Preference                                                         & 18\% (7/38)  \\
                                                                                                                                                                                                                                                                                                         & Fixed   Sizes (choose from: small, medium, and large)                   & 11\% (4/38)  \\ \hline
\multirow{4}{3in}{(3) How many hours does the platform need to   operate on a single full charge?}                                                                                                                                                                                                             & 4 - 6 hours                                                             & 35\% (13/37) \\
                                                                                                                                                                                                                                                                                                         & 2 - 4   hours                                                           & 30\% (11/37)  \\
                                                                                                                                                                                                                                                                                                         & 6+ hours                                                                & 27\% (10/37)  \\
                                                                                                                                                                                                                                                                                                         & 0 - 2   hours                                                           & 8\% (3/37)   \\ \hline
\multirow{2}{3in}{(4) Quori will feature a low-cost   holonomic base by default. What modular mobility options would you prefer?}                                                                                                                                                                              & Mobile                                                                  & 1.74            \\
                                                                                                                                                                                                                                                                                                         & Tabletop                                                                & 1.33            \\ \hline
\multirow{3}{3in}{(5) What data storage options would   you prefer?}                                                                                                                                                                                                                                           & Removable                                                               & 2.59            \\
                                                                                                                                                                                                                                                                                                         & Internal                                                                & 1.82            \\
                                                                                                                                                                                                                                                                                                         & External                                                                & 1.71            \\ \hline
\multirow{2}{3in}{(6) What computing options would you   prefer?}                                                                                                                                                                                                                                              & On-board                                                                & 72\% (26/36) \\
                                                                                                                                                                                                                                                                                                         & Off-board                                                               & 28\% (10/36) \\ \hline
\end{tabular}
\end{table}
\begin{table}[ht]
\centering
\caption{Survey \#2: Robot Sensing and Behavioral Considerations}
\label{table:survey2-sen}
\begin{tabular}{|m{4.5in}cc|}
\hline
Prompt                                                                                                                                                                                                                                                                                                                                                                                                                                                                          & Responses         & Result          \\ \hline
\multirow{4}{4.5in}{(1) By default, Quori will include several low-level software “drivers” (speech, eye gaze, gestures, proxemics, emotion, etc.) for behavior generation and recognition that are compatible with Robot Operating System (ROS). How much do you intend to modify Quori’s behavior generation and recognition systems?}                                                                                                                                        & Might modify      & 56\% (19/34) \\
                                                                                                                                                                                                                                                                                                                                                                                                                                                                                & Will   modify     & 26\% (9/34)  \\
                                                                                                                                                                                                                                                                                                                                                                                                                                                                                & Will not   modify & 18\% (6/34)  \\ & & \\\hline
\multirow{5}{4.5in}{(2) Some (but not all) commercial   (i.e., closed source, often for purchase, often with user/customer supported)   software outperforms existing non-commercial (i.e., open source, often for   free, sometimes with user support) software solutions for autonomous behavior   generation and recognition. How do you rank commercial vs. non-commercial   software? Would you use commercial software if it performs better and/or has   user/customer support?} & No                & 55\% (17/31) \\
                                                                                                                                                                                                                                                                                                                                                                                                                                                                                & Yes               & 45\% (14/31) \\ & &\\& &\\& &\\ \hline
\multirow{12}{4.5in}{(3) By default, Quori will speak   English; however, some users may need to support other spoken languages. What   spoken languages (other than English) should be supported by the platform?}                                                                                                                                                                                                                                                                     & Spanish           & 11.21           \\
                                                                                                                                                                                                                                                                                                                                                                                                                                                                                & Mandarin          & 10.67           \\
                                                                                                                                                                                                                                                                                                                                                                                                                                                                                & Japanese          & 8.92            \\
                                                                                                                                                                                                                                                                                                                                                                                                                                                                                & German            & 8.46            \\
                                                                                                                                                                                                                                                                                                                                                                                                                                                                                & Hindi             & 6.64            \\
                                                                                                                                                                                                                                                                                                                                                                                                                                                                                & French            & 6.53            \\
                                                                                                                                                                                                                                                                                                                                                                                                                                                                                & Portuguese        & 6.38            \\
                                                                                                                                                                                                                                                                                                                                                                                                                                                                                & Arabic            & 6.17            \\
                                                                                                                                                                                                                                                                                                                                                                                                                                                                                & Korean            & 5.25            \\
                                                                                                                                                                                                                                                                                                                                                                                                                                                                                & Russian           & 5.20             \\
                                                                                                                                                                                                                                                                                                                                                                                                                                                                                & Italian           & 4.30             \\
                                                                                                                                                                                                                                                                                                                                                                                                                                                                                & Bengali           & 4.10             \\ \hline
\end{tabular}
\end{table}
\begin{table}[ht]
\centering
\caption{Survey \#2: Demographics}
\label{table:survey2-dem}
\begin{tabular}{|m{2in}m{3in}c|}
\hline
Prompt                                                                                                    & Responses                                                                                                                                                                      & Result          \\ \hline
\multirow{3}{2in}{(1) How often do you personally (i.e., as an   individual) … \\ \{Weekly, Monthly, Yearly, Never\}} & … modify, but do   not publicly release, existing open-source software (i.e., software that   originated from another individual or organization outside of your   workplace)? & \{55, 60, 35, 22\}\% \\
                                                                                                          & …   modify and publicly release existing open-source software (i.e., software   that originated from another individual or organization outside of your   workplace)?          & \{10, 20, 25, 47\}\% \\
                                                                                                          & … create   and publicly release new open-source software (i.e., software that originated   from your own ideas or workplace)?                                                  & \{35, 20, 40, 31\}\% \\ \hline
\multirow{6}{2in}{(2) What is your age (in years)?}                                                             & 25-34                                                                                                                                                                          & 13\% (13/36)    \\
                                                                                                          & 35-44                                                                                                                                                                          & 10\% (10/36)    \\
                                                                                                          & 45-54                                                                                                                                                                          & 6\% (6/36)      \\
                                                                                                          & 18-24                                                                                                                                                                          & 5\% (5/36)      \\
                                                                                                          & 55-64                                                                                                                                                                          & 1\% (1/36)      \\
                                                                                                          & 65-74                                                                                                                                                                          & 1\% (1/36)      \\ \hline
\multirow{3}{2in}{(3) What is your gender?}                                                                     & Male                                                                                                                                                                           & 23\% (23/36)    \\
                                                                                                          & Female                                                                                                                                                                         & 11\% (11/36)    \\
                                                                                                          & Did not   specify                                                                                                                                                              & 2\% (2/36)      \\ \hline
\multirow{4}{2in}{(4) What is your ethnicity? (Please select all that   apply.)}                                & White/Caucasian                                                                                                                                                                & 77\% (27/35) \\
                                                                                                          & Asian or   Pacific Islander                                                                                                                                                    & 14\% (5/35)   \\
                                                                                                          & Did not   specify                                                                                                                                                              & 9\% (3/35)   \\
                                                                                                          & Hispanic   or Latino                                                                                                                                                           & 6\% (2/35)    \\ \hline
\multirow{8}{2in}{(5) How much do you know about the following topics?}                                       & Human-Robot   Interaction                                                                                                                                                      & 5.31            \\
                                                                                                          & Artificial   Intelligence                                                                                                                                                      & 5.00               \\
                                                                                                          & Robotics                                                                                                                                                                       & 4.94            \\
                                                                                                          & Psychology                                                                                                                                                                     & 4.00               \\
                                                                                                          & Signal   Processing                                                                                                                                                            & 3.46            \\
                                                                                                          & Animation   (2D/3D)                                                                                                                                                            & 3.06            \\
                                                                                                          & Anthropology                                                                                                                                                                   & 2.80             \\
                                                                                                          & Anatomy                                                                                                                                                                        & 2.63            \\ \hline
\end{tabular}
\end{table}
%

\end{document}